\documentclass{article}
\usepackage{arxiv}
\usepackage[utf8]{inputenc} 
\usepackage[T1]{fontenc}    
\usepackage{hyperref}       
\usepackage{url}            
\usepackage{booktabs}       
\usepackage{amsfonts}       
\usepackage{nicefrac}       
\usepackage{microtype}      
\usepackage{lipsum}		
\usepackage{graphicx}
\usepackage{natbib}
\usepackage{doi}
\usepackage{amsmath,amsfonts,amsthm} 
\usepackage{amsbsy}  
\usepackage{amssymb}  
\usepackage{subcaption}  
\usepackage{float}  

\title{Word embeddings for topic modeling: an application to the estimation of the economic policy uncertainty index}

\date{September 15, 2021}	

\author{ Hairo Ulises Miranda Belmonte \\
	Centro de Investigación en Matemáticas (Cimat) \\
	Monterrey, 66628, Mexico\\
	\texttt{hairo.miranda@cimat.mx} \\
	\And
	Victor Muñiz-Sánchez\thanks{Corresponding author. Please, if you require to quote this working progress, request it to the corresponding author. } \\
	Centro de Investigación en Matemáticas (Cimat) \\
	Monterrey, 66628, Mexico\\
	\texttt{victor\_m@cimat.mx} \\
	
	\And
	Francisco Corona \\
	Instituto Nacional de Estadística y Geografía (INEGI) \\
	Mexico City, Mexico\\
	\texttt{franciscoj.corona@inegi.org.mx}
}

\renewcommand{\shorttitle}{\textit{arXiv} Template}

\hypersetup{
pdftitle={A template for the arxiv style},
pdfsubject={q-bio.NC, q-bio.QM},
pdfauthor={David S.~Hippocampus, Elias D.~Striatum},
pdfkeywords={First keyword, Second keyword, More},
}

\begin{document}
\maketitle

\begin{abstract}
Quantification of economic uncertainty is a key concept for the prediction of macro economic variables such as gross domestic product (GDP), and it becomes particularly relevant on real-time or short-time predictions methodologies, such as nowcasting \citep{banbura2010, banbura2013}, where it is required a large amount of time series data, commonly with different structures and frequencies. Most of the data comes from the official agencies statistics and non-public institutions, however, relying our estimates in just the traditional data mentioned before, have some disadvantages. One of them is that economic uncertainty could not be represented or measured in a proper way based solely in financial or macroeconomic data (\cite{Baker2016}), another one, is that they are susceptible to lack of information due to extraordinary events, such as the current COVID-19 pandemic. For these reasons, it is very common nowadays to use some \emph{non-traditional} data from different sources, such as social networks or digital newspapers, in addition to the traditional data from official sources. The economic policy uncertainty (EPU) index (\cite{Baker2016}), is the most used newspaper-based indicator to quantify the uncertainty, and is based on topic modeling of newspapers. In this paper, we propose a methodology to estimate the EPU index, which incorporates a fast and efficient method for topic modeling of digital news based on semantic clustering with word embeddings, allowing to update the index in real-time, which is a drawback with another proposals that use computationally intensive methods for topic modeling, such as Latent Dirichlet Allocation (LDA). We show that our proposal allow us to update the index and significantly reduces the time required for new document assignation into topics.
\end{abstract}

\keywords{EPU-index \and Word-embeddings \and Fuzzy k-means}

\section{Introduction}
Uncertainty trigger affects economic activity  on both the demand and supply side \citep{stock2012}. On the supply side, a reduction occurs in the willingness of companies to hire more workers,  thereby increasing the unemployment rate and contracting consumption. On the demand side, investors reduce their expectations in the short-run, inducing changes in the financial market and the exchange rate. These effects produce instability in the economic activity growth rate, measured by the Gross Domestic Product \citep{Baker2020}, forcing policymakers to react through fiscal and monetary policy.

Uncertainty around the world has been increased by the recent COVID-19 pandemic, and according to the organisation for economic co-operation and development (OECD) latest Economic Outlook from 2020, the world economic output forecast is expected to plummet $7.6\%$ this year and unemployment in OECD economies is expected to be more than double. COVID-19 has made  policymakers implement social distancing, business lockdowns and other strategies to stop the propagation of the virus. As a result, the massive spike in uncertainty and the inability to measure it in real time makes it almost impossible to quantify the uncertainty in individual countries. We can find in the literature different methods to measure economic uncertainty. In general,  we can divide the approaches into two categories: those related to structured data (financial or macroeconomic)  combined with statistical and econometric methods, and those related to unstructured data, such as text from newspapers or social media, generally combined with machine learning algorithms.

In recent years, uncertainty in the economy has been associated with the financial market \citep{Bloom2009}. Hence, in the first approach,  uncertainty is quantified by the volatility of the stock market, for example, the index VXO  based on the S\&P 100, or the VIX based on the S\&P 500, both created by the Chicago Board of Options Exchange. Some studies use the decomposition of market volatility from external  shocks and financial firms returns to measure uncertainty \citep{cambel2001, bloom2007, Gilch2014, carriero2018, lahir2010}. Techniques used to quantify these effects include  structural vector autoregression (SVAR) models \citep{Gilch2014}, conditional variance or generalized autoregressive conditional heteroskedasticity (GARCH) models \citep{Ferrera2014}, and stochastic volatility  \citep{carriero2018}. In a micro or macro economic context, some researchers make use of surveys of firms to measure uncertainty \citep{Ezgi2017, Jo2015, Rossi2015, Scoti2016, bach2013}.

Recently, many studies have applied the second approach, i.e., newspaper-based measures of uncertainty, such as the EPU index \citep{Baker2016}. The EPU index quantifies the dynamics of newspapers based on their latent topics, and it is estimated by counting the number of articles in a given time. To create the indicator, Baker used the number of news associated with economic uncertainty and manually classifies topics into eight subcategories: fiscal, monetary policy, healthcare, national security, regulation, sovereign debt and currency crisis, entitlement programs and trade.

Baker's index measures uncertainty in real time but is constructed manually. To address this issue, \cite{Azqueta2017}  improves EPU index via automation of the topic extraction step. In Azqueta's research, EPU index is replicated for the US economy using three newspaper sources: USA today, The Washington Post, and The New York Times. The news were obtained from a non-free database with extensive media article coverage. The newspapers were selected based on those with the words economy and uncertainty. After applying some text preprocessing techniques, Azqueta applied latent Dirichlet allocation (LDA) to classify the words into topics and the topics into documents. Next, he calculated the index by standardizing the topic series by the number of times the word "today" appear in a month. Finally, he took the topic series  associated with Baker's subcategories and aggregated them to normalize the results into specific values\footnote{Mean 100 and standard deviation of one. }.

The usefulness of the index has been increasing, and is currently available in other countries (\cite{Azqueta20201, Azqueta20202}). In addition, \cite{Baker2020} propose  the EPU index to estimate the uncertainty of the recent COVID-19 pandemic and its influence on the world economy. 

One advantage of the EPU index is the ability to explain the uncertainty associated with economic subcategories and the source of daily periodicity. We can add  topics to replicate uncertainty indicators to quantify financial data, such as the VIX stock market index \citep{Azqueta20201, Azqueta20202}. 
On the other hand, \cite{Ahir2020} noted some disadvantages of the EPU index. They mention that methods based on text-searching newspapers are limited to the set of mostly advanced economies. \cite{Azqueta20201} remark that LDA is a time-consuming algorithm, making it difficult to update the EPU index in real-time.

 \cite{Ahir2020}  address some of the EPU limitations by introducing the world uncertainty index (WUI). This index is a new quarterly measure of uncertainty that covers 143 countries\footnote{Countries with a population of at least 2 million.}. They use country reports from the economist intelligence unit (EIU)\footnote{The EIU is a business intelligence company that provides quarterly country reports that cover economics, policies, and political topics.} to estimate the index.  The following steps summarize the process used to estimate the WUI index:  compile the quarterly announcements; count the number of times the word “uncertainty”  is found in the document; and normalize based on the total number of words in each report. Some of the limitations of the EPU index are solved by the WUI, but as in \cite{Baker2016}, the index is constructed manually. 
 
 In this paper, we propose a novel approach to estimate the subcategories that compose the EPU index. The method consists of semantic clustering using word embeddings and fuzzy k-means to approximate the LDA mechanism. With this approach, it is easier to measure the uncertainty in real-time and update the index, thereby reducing the computational time required to estimate the latent topics and the EPU index. Additionally, we create our own corpus for US and Mexico newspapers and estimate Mexico's EPU index for a longer periodicity than that in \cite{Baker2016}  using  the \cite{Azqueta2017}  machine learning algorithm approach. 
 
The paper is structured as follows. In Section \ref{sec:background}, we give a background on word embedding and topic models, which is needed in order to understand our proposal. In Section \ref{sec:methodology}, we present the methodology we propose to estimate the EPU index, including a description of the corpus we created to construct the index. In Section \ref{sec:results} we show the results we obtained for both US and Mexico corpus, where we include a comparison of topic modeling with LDA and our proposal, the EPU subcategories we obtained and our estimation of the index itself; also, we present a \emph{complement index}, consisting on the combination of some subcategories we obtained with our proposal. Finally, in Section \ref{sec:conclusions}, we present our conclusions and future work.

\section{Background on word embeddings and topic models}
\label{sec:background}

Topic modeling is an important task in natural language processing (NLP). Although it was initially considered as part of information retrieval, topic modeling is currently part of many applications in text mining and related areas, such as text classification, automatic summarization and novelty detection.
Broadly speaking, a topic is a latent structure that can summarize semantic concepts such as the meaning or general idea of a text or sentence. 
Topic modeling is thus a statistical representation of the topics within a text corpus. Two key ingredients are needed in topic modeling: a text representation and a statistical model to extract topics and infer a probabilistic distribution of documents among the topics. In the next subsections, we will address both of these factors.

\subsection{Word and document representations}

Different approaches for topic modeling are highly dependent on the vector space representation of texts and documents. Most topic models use the vector space model \citep{salton1975}, also known as bag of words (BOW) representation, where the vector space representation is given by a term-frequency matrix, optionally with a weighted scheme, such as  term frequency-inverse document frequency (TF-IDF, \cite{salton_tfidf88}), where documents are vectors with lengths equal to the size of the vocabulary of the corpus. This representation is highly dimensional and sparse (high number of zeros), which can lead to some computational issues. Certainly, BOW and TF-IDF are common approaches to represent words and documents in a vector space for many NLP tasks where similarities must be defined. In BOW, documents are considered collections of words without any additional properties, such as sequential order or grammar.

In recent years, the convenience of using dense vector representations of a predefined length, instead of sparse vectors, for a wide variety of NLP tasks has been noted. These representations are called embeddings. In addition to computational advantages, word and document embeddings enable us to infer semantic properties by introducing a language model (LM) to obtain them. Models that assign probabilities to sequences of words are called language models. 

Let $w_1,w_2,\ldots,w_T$ be a sequence of words. A language model obtains the probabilities of single words $P(w_t)= P(w_t|w_1^{t-1})$ in the sequence and the  probability to obtain a sequence of $n$ words: 
\begin{equation}\label{eq:lm1}
    P(w_1^n)=\displaystyle \prod_{t=1}^n P(w_t|w_1^{t-1}).
\end{equation}

The latter is very difficult to obtain as the sequence increases in length, and a simplification is given by the $n-$gram model, which approximates the probability on the right side of (\ref{eq:lm1}) as $P(w_t|w_1^{t-1}) \approx P\left(w_t|w_{t-N+1}^{t-1}\right)$, defining a context based on only the $N$ previous words. When $N=1$ (bigram model), we obtain the Markov model.

There are many approaches to compute these probabilities, but one of particular interest is the neural LM (\cite{bengio2003}), which uses a feedforward neural network. A feedforward neural LM is a standard feedforward network that takes as input at time $t$, a representation (such as one-hot encoding) of some previous words ($w_{t-1},w_{t-2}, \ldots$)  and outputs a probability distribution over possible next words. In this LM, there is a vector $\mathbf{v}\in \mathbb{R}^d$ of a given dimensionality $d$ associated with every word in the vocabulary of the corpus, and the joint probability of ($w_1,w_2,\ldots,w_T$) is given in terms of $\mathbf{v}_1,\mathbf{v}_2,\ldots,\mathbf{v}_T$. Then, the embeddings are given by the vectors $\mathbf{v}$, which are the weights of the neural network used to learn the probability of words given a context of words.

There have been many proposals to obtain word and document embeddings based on the neural LM of Bengio et al. Perhaps the most popular model for word embeddings is word2vec \citep{word2vec_mikolov}, which offers a computational simplification of the neural LM by removing the nonlinear hidden layer in the architecture proposed by Bengio et al. word2vec implements two algorithms, a continuous BOW model (CBOW) and skip-gram with negative sampling (SGNS) model, and can recover interesting semantic relationships between words based on the context in which they appear, such as analogies. Another model for word embeddings, which we use in our proposal, is GloVe (\cite{pennington2014glove}). GloVe combines two methods used to obtain vector representations of words: global matrix factorization and local context window. Global matrix factorization is a method used to solve the high-dimensionality problem that results from working with a matrix of counts, such as BOW, by generating low-dimensional vector representations of words\footnote{Latent semantic analysis (LSA) is an example of a matrix factorization technique applied to BOW}. On the other hand, local context window methods, such as word2vec, can capture complex patterns of words, such as similarities, and are scalable with corpus size. GloVe takes advantage from both concepts by working with a word-to-word co-occurrence matrix (a matrix of counts similar to BOW) that captures the  global co-occurrence statistics of the corpus and produces a semantic vector space with a meaningful structure. From a statistical perspective, GloVe can be viewed as a log-bilinear regression model that learns word representations, and the authors show that this proposal outperforms other embedding models on tasks such as word analogy and word similarity.

\subsection{Topic models}

As stated previously, topic models can be analyzed according to the vector space representation they use. Models using BOW or TF-IDF representations can be classified as matrix factorization methods and generative probabilistic methods. Because these representations depend on the vocabulary of the corpus, they are generally high dimensional, and models based on matrix factorization aim to solve this problem by applying dimension-reduction techniques to the term-frequency matrix. The most popular approaches are latent semantic indexing (also known as LSA, \citep{Landauer1998}), nonnegative matrix factorization \citep{lee-NIPS2001}, and probabilistic latent semantic analysis (PLSA, \citep{Hofmann1999}). Perhaps the most popular topic modeling method is LDA, proposed by \cite{lda2003}, which is a generative probabilistic method that assumes that documents are generated by a probabilistic distribution over topics and that topics are generated by a distribution over words. 

\subsubsection{Matrix factorization methods}

What is observable in BOW is the word frequency over documents, and topics are considered hidden or latent. LSA is a method to infer these latent structures using matrix factorization on the BOW or TF-IDF matrix. Let $\mathbf{A}_{m\times n}$ be the rank $r$ term-document matrix, with $m$ rows (terms) and $n$ columns (documents). In LSA, we use a rank $k<r$ approximation based on a truncated SVD factorization, $\mathbf{A}\approx\mathbf{A}^{(k)}=\mathbf{U}^{(k)}\boldsymbol{\Sigma}^{(k)}\mathbf{V}^{(k)'} =\sum_{i=1}^k \sigma_i\mathbf{u}_i\mathbf{v}_i'$, where $\boldsymbol{\Sigma}^{(k)}$ is a diagonal matrix with the largest $k$ eigenvalues of $\mathbf{A}$ and $\mathbf{U}^{(k)}, \mathbf{V}^{(k)}$ contains the corresponding left and right eigenvectors, defining an orthonormal basis for the column and row spaces, respectively. The intuition behind this factorization in LSA is that it retains the most important information of $\mathbf{A}$, which in this context, means that it captures the conceptual and semantic relationship among words and documents. Generally, $k<< r$  to obtain a dense low-dimensional representation of the term-document matrix, and, for topic modeling, $\mathbf{U}^{(k)}$ represents the word assignment to topics, $\mathbf{V}^{(k)}$ represents the topic distribution across documents and $\boldsymbol{\Sigma}^{(k)}$ represents topic importance. The choice of the number of topics $k$ is critical because it must be sufficiently large to fit all the real structure in the data but sufficiently small to avoid fitting the sampling error or unimportant details. A general approach to select $k$ is to use the topic coherence measure \citep{chang_NIPS2009, newman2010}. One drawback of LSA is the lack of interpretability due to mixed signs in the basis vectors of $\mathbf{U}$ and $\mathbf{V}$. Nonnegative matrix factorization (NMF) solves this problem for nonnegative data, such as BOW representations. For a matrix $\mathbf{A}$ with nonnegative entries, NMF uses a low-rank approximation with nonnegative factors $\mathbf{A}\approx\mathbf{A}^{(k)}=\mathbf{W}^{(k)} \mathbf{H}^{(k)}$, where $\mathbf{W}^{(k)}, \mathbf{H}^{(k)} \geq 0$. Different algorithms have been proposed to obtain the factors \citep{cichocki2009, lee-NIPS2001} and solve the nonlinear optimization problem defined by

\begin{equation}
    \label{eq:nmf1}
    \begin{aligned}
        \displaystyle \min_{\mathbf{W}^{(k)}, \mathbf{H}^{(k)}} \quad & f(\mathbf{A};\mathbf{W}^{(k)}, \mathbf{H}^{(k)}) \\
        \textrm{s.t.} \quad & \mathbf{W}^{(k)}, \mathbf{H}^{(k)} \geq 0,
    \end{aligned}
\end{equation}

where $f(\cdot)$ is an appropriate cost function, such as the mean squared error with the Frobenius norm or Kullback-Leibler divergence. Problem (\ref{eq:nmf1}) can be adapted to obtain a sparse representation on the factors, facilitating the interpretability of the solution. NMF has good scalability as $k$, $m$ and $n$ increase, and in computational terms, can be better than LSA; however, convergence can be guaranteed only to local minima for some algorithms.

\subsubsection{Latent Dirichlet Allocation}

\begin{figure}[H]
  \centering
  \includegraphics[scale=.4]{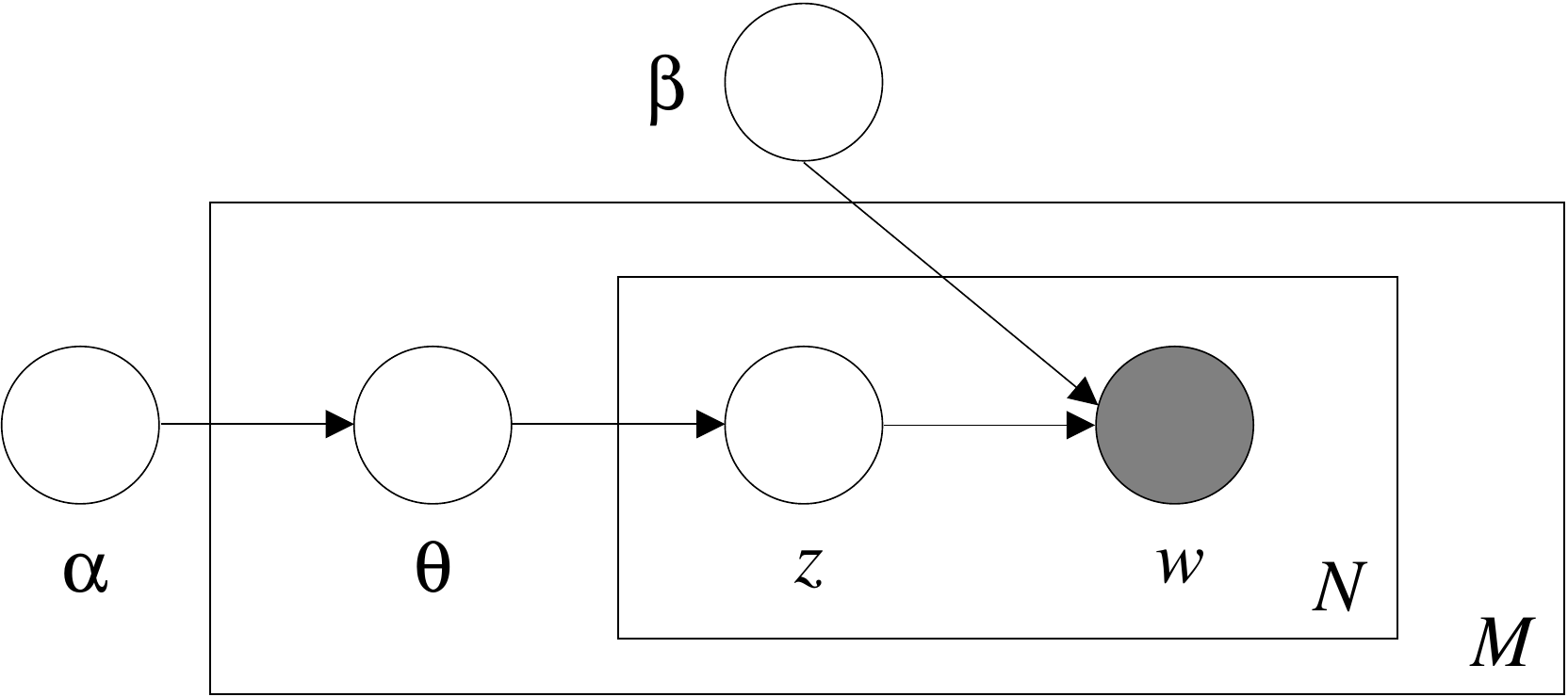}
   \caption{Graphical representation of the LDA hierarchical Bayesian model (\cite{lda2003}).}
    \label{fig:lda_model}
\end{figure}

LDA is described in \cite{lda2003} as "a generative probabilistic model for collections of discrete data such as text corpora". The intuition of this model is that, given a corpus of documents with a BOW representation, each document is a mixture of corpus-wide topics, each topic is represented as a distribution over words and each word is drawn from one of those topics. Consider a corpus with $M$ documents, where each document $\mathbf{w}$ is a sequence of $N$ words and each word belongs to a finite vocabulary of size $V$. LDA is a three-level hierarchical Bayesian model represented graphically with plate notation in Figure \ref{fig:lda_model} that represents the generative process of documents, where, for each document $\mathbf{w}$, we choose a topic proportion $\theta \sim \text{Dir}(\alpha)$, and for each of the $N$ words $w_n$, we choose a topic $z_n\sim \text{Multinomial}(\theta)$ and a word $w_n$ from $p(w_n|z_n,\beta)$, which is a multinomial probability function conditioned on topic $z_n$. 

Given this model for the generative process for documents, the goals is to compute $p(\text{topics, proportions,assignments}|\text{documents})$, which can be formally expressed as a joint distribution with the following parameterization:

\begin{displaymath}
    p(\theta,\mathbf{z},\mathbf{w}|\alpha,\beta)=
    p(\theta|\alpha)\prod_{n=1}^N p(z_n|\theta)p(w_n|z_n,\beta).
\end{displaymath}
For the indexes $n=1,\ldots,N$ (words), $v=1,\ldots,V$ (vocabulary), and $m=1\ldots,M$ (documents) and considering a fixed number $k$ of topics, in this model, $z_{n,m}$  is the per-word topic assignment, $\theta_m$ is the per-document topic proportion and $\beta_{k,v}$ is the per-word probability, such that $\beta_{k,v}=p(w^v=1|z^k=1)$. In this setting, $\theta$ is a $k-$dimensional Dirichlet random variable parameterized by $\alpha \in \mathbb{R}^k$, which controls the mean shape and sparsity of $\theta$. The Dirichlet distribution is an exponential family distribution over the simplex (positive vector summing to one) and is very convenient because it is conjugate to the multinomial distribution. Details about the prior distributions for $\alpha$ and $\beta$ can be found in \cite{lda2003}.

Inference and parameter estimation in LDA is complex and computationally demanding. The inference step consists of computing the posterior distribution of the latent variables given a document $p(\alpha,\mathbf{z}|\mathbf{w},\alpha,\beta)= \frac{p(\theta,\mathbf{z},\mathbf{w}|\alpha,\beta)}{p(\mathbf{w}|\alpha,\beta)}$; however, as noted by Blei et.al., this distribution is intractable for exact inference and should be approximated via techniques such as Markov chain Monte Carlo (MCMC), Laplace approximation or variational approximation. The latter is used by Blei et al. Parameter estimation for $\alpha$ and $\beta$ is performed via a variational EM algorithm.

Some extensions to LDA have been proposed. \cite{Blei2005} proposed the correlated topic model (CTM), where the topic proportions exhibit correlation from a logistic normal distribution. \cite{Yan2013} proposed  a topic model for short text named the biterm topic model (BTM), which extract topics by directly modeling the generation of word co-occurrence patterns. Additionally, \cite{Yin2014} used the Dirichlet multinomial mixture (DMM), which is a variant of LDA that uses unigram mixtures and assumes that each text is sampled from just one topic. \cite{Liyan2018} used LDA with a TF-IDF weight scheme for topic trend detection (TTD) applied to the Sina Microblog collection of short texts. Moreover, \cite{Pagare2020} proposed an LDA extension called trendy topic detection and trendy community detection (T-ToCODE), which identifies hot topics and communities associated with every popular topic on Twitter. Essentially, the T-ToCODE considers different features, such as  hashtags, comment count, retweet count, and likes, for a particular post to identify if the topic is trendy.

\subsubsection{Other approaches}

There are other approaches for topic modeling that have been proposed in the literature, based on different word and document representations and different mechanisms to identify topics and assign documents to them. In this subsection, we will mention some of them which are related in some way to our proposal for topic modeling.

\cite{Bafna2016} used the TF-IDF representation together with fuzzy $k-$means \citep{dunn73} to obtain topics and assign documents into those topics for general-purpose corpus. \cite{Zhu2019} proposed the TA TF-IDF algorithm, which is a refined TF-IDF representation to overcome its shortcomings in explaining the importance of hot-terms (topics) by adding time distribution information and user attention in digital news, where the time distribution is used to identify the active period of hot-terms. The user attention is related to the number of news hits and the number of user participation, such as news comments. Then, they  used $k-$-means on this representation to identify the relevant topics. \cite{Li-Qiang2015} proposed Topic2Vec, an extension of LDA using word2vec embeddings, to learn distributed topic representations together with word representations. Topic2vec consider CBOW and Skip-gram: CBOW predicts the word and topic  based on a context window, while Skip-gram predicts the surrounding words given the target word and its topic. For training, the model requires a set of word-topic sequences of a document, and the word's topic is inferred via LDA. \cite{Li2018} proposed partitioned Word2Vec-LDA (PW-LDA) topic model, which is particularly useful in short texts such as academic abstracts. The model aims to reduce the sparsity of short texts and makes it possible to detect topics in narrow-domain corpus with a combination of purpose sentences extraction method, high-frequency words removal and LDA with different Word2vec models. The topic embedding is a weighted sum of the different representations obtained. In other research, \cite{Kim2020} proposed a topic model called Word2vec-based latent semantic analysis (word2vec-LSA) that makes use of Word2vec, a contextual word embedding algorithm and spherical $k-$means clustering. The approach was applied to the annual trends in blockchain.

Clustering in the embedding space has also been used, mainly with texts from twitter and using word2vec and FastText \citep{fasttext_bojanowski-etal-2017} embeddings with $k-$means algorithm \citep{Vargas2019a, Vargas2019b, Vargas2019c}. Additionally,  \cite{Chan2017} used  Word2Vec, FastText, and Doc2Vec (\cite{doc2vec}) to generate a semantic space, and proposed a clustering-based language model by using Brown \citep{brown-etal-1992-class} and $k-$means algorithms to obtain text readability prediction and sentence matching based on semantic relatedness, trained in the Common Core Standards and Wiki-SimpleWiki corpus. Recently, \cite{Meghana2020} used word2vec skip-gram model and k-means clustering to identify semantic similarities from a set of articles related to different domains and newspapers that contain many categories. 

Additionally, some authors have proposed different approaches to detect topics in texts, mainly from social media, which does not depend on a particular word or document representation. For example, \cite{Choi2019} implemented a  method to detect emerging topics on Twitter based on high utility pattern mining (HUPM) instead of other methods that consider only the frequencies of words and not their utility  to detect the topic based on the growth rate in appearance frequency. The work of \cite{Sun2020} uses the conversational structure-aware topic model (CSATM) for short text topic detection. This model considers the discussion thread tree structure and proposes a popularity metric to quantify the number of responses to a given comment and the transitivity concept to characterize topic dependency among nodes in a nested discussion thread. 

\section{Methodology}
\label{sec:methodology}

In this section, we present our approach for estimating the EPU index, which consists of a topic model based on fuzzy clustering in the semantic space spanned by word embeddings, as a computationally efficient alternative to LDA. Finding clusters in the embeddings space means that the words in these groups are semantically related. We use fuzzy $k-$means (\cite{dunn73}, \cite{kaufman90}) to infer the probability distribution of the words belonging to specific topics;  then, we assign documents to topics with an appropriate dissimilarity measure defined in the semantic space.

Consider a set of word embeddings $\mathbf{v}_i,\ldots,\mathbf{v}_n$ and a fixed number of topics $K$. Fuzzy $k-$means (also called \emph{soft} $k-$means) is a combinatorial algorithm to find a set of cluster centers $C=(\mathbf{c}_1,\ldots,\mathbf{c}_K)$ by solving the following optimization problem:

\begin{equation}
    \label{eq:fuzzyK}
    \begin{aligned}
        \displaystyle \min_{C} \quad & \sum_{k=1}^K\sum_{i=1}^n p_k(\mathbf{v}_i)^q\Vert \mathbf{v}_i-\mathbf{c}_k \Vert^2 \\
        \textrm{s.t.} \quad & \sum_{k=1}^K p_k(\mathbf{v}_i) =1, \quad p_k(\mathbf{v}_i)\geq 0.\\
    \end{aligned}
\end{equation}
In fuzzy clustering, each data point (i.e., each word represented by its embedding) has a probability of belonging to each of the $k$ clusters (i.e., topics). This probability is given by $p_k(\mathbf{v}_i)\in[0,1]$ in equation (\ref{eq:fuzzyK}), which is computed by
\begin{equation}
    \label{eq:fuzzyK-p}
    p_k(\mathbf{v}_i) =  \frac{1}{\displaystyle \sum_{j=1}^K 
    \left(\frac{\Vert \mathbf{v}_i-\mathbf{c}_k \Vert}
    {\Vert \mathbf{v}_i-\mathbf{c}_j \Vert}\right)^{2/(q-1)}},
\end{equation}
where the superscript $q$ is a fuzzy constant that controls the level of clustering fuzziness.

Similar to LDA and given a corpus of documents, we assume there is a distribution of words into topics and topics into documents. 

Once we obtain word embeddings learned from a given corpus, and a cluster model, cluster allocation of new documents can be done dynamically and fast because we just need to obtain the vector representation of the new document based on the semantic space learned before. Then, with a similarity measure, such as the cosine distance, we can determine to which cluster (i.e., topic) it belongs. As a consequence, our proposal will reduce the time and computational effort required for topic allocation because it is not necessary to fit the model again. 

Finally, to obtain the EPU index, we use the procedure and standardization method followed by \cite{Baker2016}. 

\begin{figure}[H]
  \centering
  \includegraphics[width=.7\linewidth]{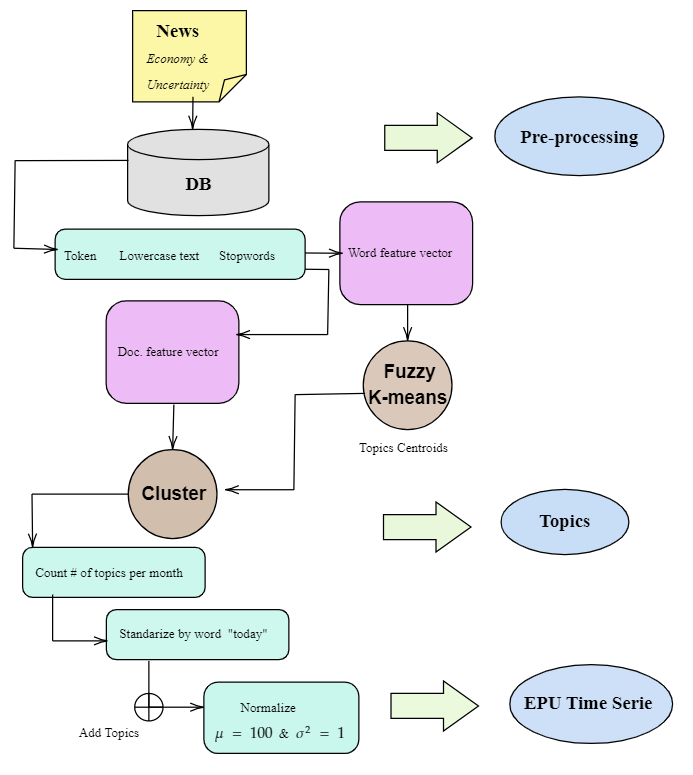}
   \caption{Proposed method}
    \label{fig:HeuristicMethod}
\end{figure}

The proposed methodology is depicted in Figure \ref{fig:HeuristicMethod} and is described below.

\begin{enumerate}
\item Compute embeddings with word2vec or Glove for words and documents in the news corpus. For documents, we take the average of all word embeddings of the document. 
\item Fit a clustering model using fuzzy $k-$means in the word embeddings representation.
\item Find the topic distribution for each document. For practical purposes, we need to assign one topic for each document. This can be done by taking the topic with the maximum probability or by using the nearest centroid (topic) to the document embedding based on cosine distance.
\item Count the number of documents assigned to each topic per month and standardize the result by the number of newspapers with the word "today"\footnote{For Mexican news corpus,  use the word "Hoy"}.
\item Generate the EPU index by summing the topics vectors from step 4, and normalize the result to a mean of $ 100 $ and a standard deviation of 1.
\end{enumerate}

\subsection{Corpus and preprocessing}

We used two corpora. For the US, we used news from \emph{The New York Times} website, and for Mexico, we used news from the websites of \emph{El Norte}, \emph{Reforma} and \emph{Mural}. As in \cite{Azqueta2017}, we filter news by the words "economy" and "uncertainty" ("economía" and "incertidumbre", for Mexico). For the US, we retrieve news from January 1990 to March 2020 for a total of $11,611$ news documents and a vocabulary size of $55,625$. For Mexico, we used the same period of time \footnote{Because these Mexican newspapers belong to the same editorial group, they publish the same news sometimes, so in order to avoid duplicates, from January 1990 to October 1993, we just used \emph{El Norte}, from November 1993 to January 1998 we used \emph{Reforma} and \emph{El Norte}, and from February 1998 to March 2020 we used all three newspapers.}; the size of the corpus is $19,149$ news documents with a vocabulary size of $59,200$. 

Preprocessing of the corpus consisted of removing stopwords and noninformative words such as abbreviations or acronyms, transforming text to lowercase, and removing numbers and special characters. After we obtained word embeddings, and following the suggestion of \cite{Abdulateef2019}, who states that text clustering can be used to eliminate redundancy and to detect semantically correlated sentences, we performed $k-$means clustering analysis to detect clusters of words that are considered noise according to their semantic relation and that do not provide relevant information to our analysis.

\subsection{Word embeddings} 
 
One problem we face in many NLP tasks when we have a small- to medium-size training corpus (as is our case), is that the assumption that training and future data must be in the same feature space and have the same distribution does not hold (\cite{pan_tf2010}). In our case, it means that we could not represent semantic properties of some words and documents because we did not see them or the specific contexts in which they appear. A common  approach to solve this problem is \emph{transfer learning}, also known as \emph{domain adaptation}. Transfer learning refers to the situation where what has been learned in one domain is exploited to improve generalization in another domain (\cite{goodfellow-et-al-2016}). In our case, we used pretrained models for word embeddings both for word2vec-CBOW and GloVe. For news in the Spanish language, we used models pretrained with the Spanish Billion Word corpus (\cite{SpanishBillionWord}), containing a vocabulary of $1,000,653$ words ($855,380$ for GloVe). For news in the English language, we used a word2vec-CBOW model pretrained with the Google News corpus (open source), with a vocabulary of almost 100 billion words, and a GloVe model pretrained with a corpus of Wikipedia documents, with a vocabulary size of approximately 400 million words. In all cases, we used embedding vectors of length $300$.

\subsection{Number of topics}

In our proposal, we need to define the number of topics, i.e., clusters. Given a set of cluster assignments, there are many proposals to assess the fit of the assignments to choose the best one. The silhouette plot (\cite{Rousseeuw1987}) and gap statistic (\cite{Tibishirani_gap}) are among the most popular; however, in our case, there is a difficulty in applying these approaches due to the dimensionality of the word embeddings and the lexical and semantic richness of the text corpora. As noted in \cite{Vargas2019a} and \cite{Vargas2019b}, if a low number of latent topics is provided, each latent topic may contain several real topics; on the other hand, if a high number of latent topics is provided, many latent topics may refer to the same real topic. \cite{Baker2016} proposed to use 8 topics related to the categories fiscal, monetary policy, healthcare, national security, regulation, sovereign debt and currency crisis, entitlement programs and trade. \cite{Azqueta2017} tried to recover those topics for the EPU index starting with 30 topics selected with a Bayesian criterion and then choosing those which are similar or seems to belong to those defined by Baker. In our case, and in order to compare the results, we used 8 topics that we named according to the words they include, following the suggestion of \cite{Azqueta2017}, as shown in Table \ref{tab:tabla1}. 

\begin{table}[H]
\footnotesize
\centering
\begin{tabular}{llllll}
    \toprule
     \textbf{EPU subcategory}  & LDA Topic  & Top Keywords   \\
   \midrule
    Fiscal Policy & Fiscal Policy & {tax, budget, cut ...} \\
        Monetary Policy & Monetary Policy & {fed, economic, rate ...} \\
        Healthcare & Healthcare & {health, airline, medic ...} \\
        National Security & Conflict, Russia, Immigration & {iraq, war, russia, refuge, immigrant..} \\
        Regulation & Law, Energy, Stock, Financial & {court, law, plant, water, stock ..} \\
        Sovereign debt & Financial crisis & {bank, loan, financi...} \\
       Entitlement Programs & Education & {school, student, college, univers...} \\        
       Trade Policy & Trade & {(china, trade, japan...} \\   
       \bottomrule
\end{tabular}
\caption{EPU subcategories matched by topics \citep{Baker2016}. Taken from \cite{Azqueta2017}}.
\label{tab:tabla1}
\end{table}

\section{Results}
\label{sec:results}

\subsection{Topic modeling comparison}

In this section, we present a qualitative comparison of topic models with LDA and our proposal based on fuzzy k-means with word embeddings, as described in Section \ref{sec:methodology}. Although it is not our aim to reproduce the same results of LDA, we present this comparison to see how sensitive  the topic distribution is to the word representations we used and what the similarities and differences are with a parametric and complex model such as LDA. In our proposal, we present the results we obtained using word2vec, word2vec with transfer learning and wieghts pretrained by Google (word2vec-pretrained), and GloVe. We present results for US and Mexican newspapers. 

\subsubsection{US corpus}
\label{sec:topic_uscorpus}

\begin{figure}[H]
    \begin{minipage}[b]{0.5\linewidth}
    \centering
  \subcaption{LDA words} 
  \centering
  \includegraphics[width=.5\linewidth]{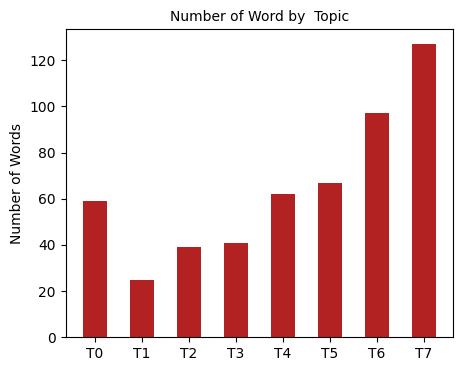}
    \vspace{1ex}
  \end{minipage}
    \begin{minipage}[b]{0.5\linewidth}
    \centering
    \subcaption{LDA news}
  \centering
  \includegraphics[width=.5\linewidth]{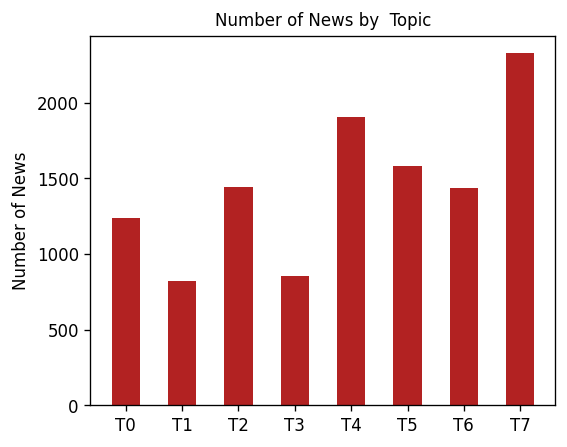}
    \vspace{1ex}
  \end{minipage}
  
    \begin{minipage}[b]{0.5\linewidth}
    \centering
  \subcaption{word2vec words} 
  \centering
  \includegraphics[width=.5\linewidth]{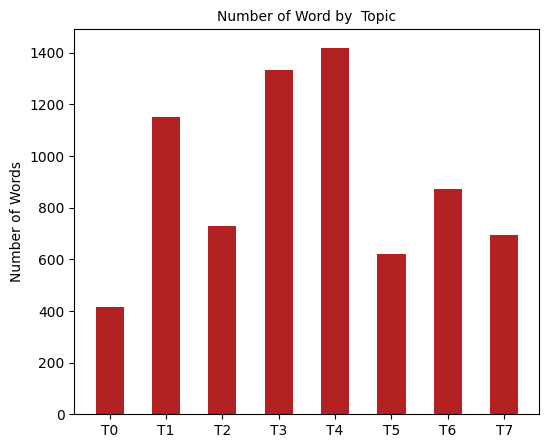}  
    \vspace{1ex}
  \end{minipage}
    \begin{minipage}[b]{0.5\linewidth}
    \centering
    \subcaption{word2vec news}
  \centering
  \includegraphics[width=.5\linewidth]{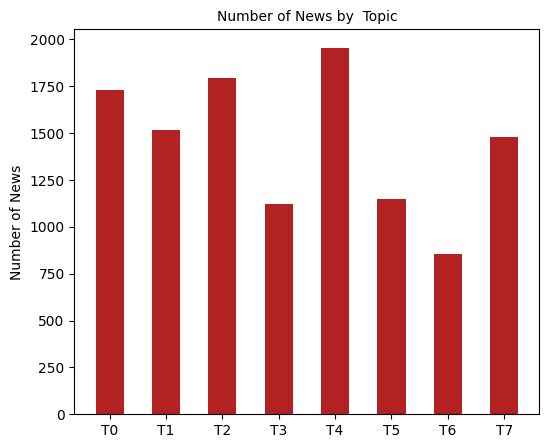}
    \vspace{1ex}
  \end{minipage}
  
  \begin{minipage}[b]{0.5\linewidth}
    \centering
    \subcaption{word2vec-pretrained words}
  \centering
  \includegraphics[width=.5\linewidth]{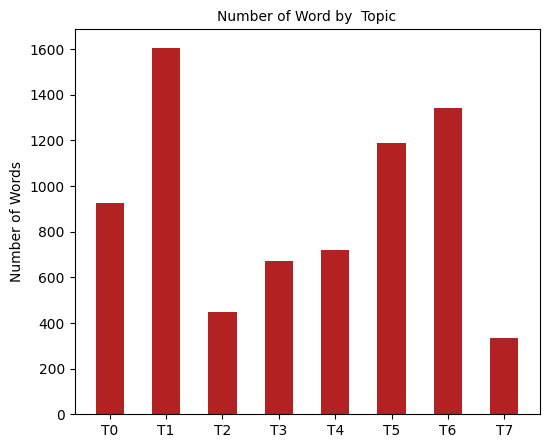}
  \centering \vspace{1ex}
  \end{minipage} 
  \begin{minipage}[b]{0.5\linewidth}
    \centering
    \subcaption{word2vec-pretrained news}
  \centering
  \includegraphics[width=.5\linewidth]{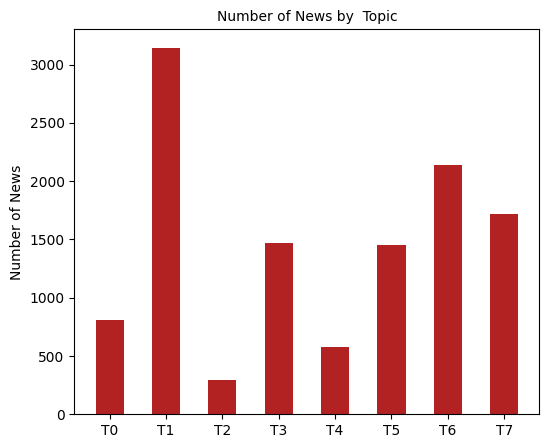}
  \centering \vspace{1ex}
  \end{minipage}
    \begin{minipage}[b]{0.5\linewidth}
    \centering
    \subcaption{GloVe words}
  \centering
  \includegraphics[width=.5\linewidth]{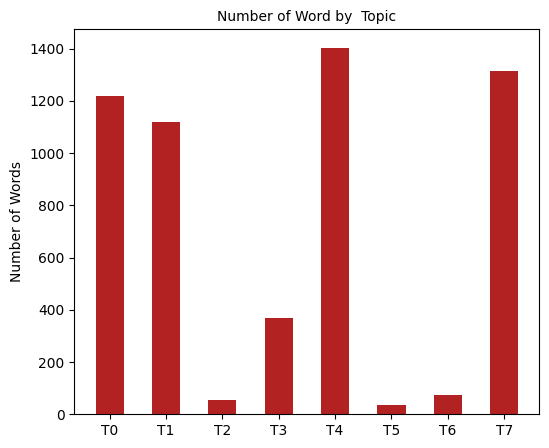}
  \centering \vspace{1ex}
  \end{minipage} 
  \begin{minipage}[b]{0.5\linewidth}
    \centering
    \subcaption{GloVe news}
  \centering
  \includegraphics[width=.5\linewidth]{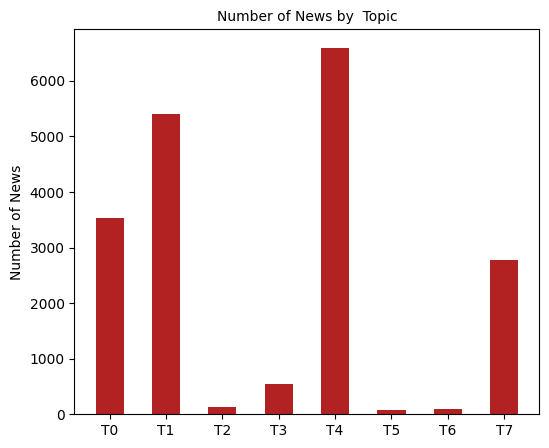}
  \centering \vspace{1ex}
  \end{minipage}
    \caption{Frequency of Words and Documents by Topic (US)}
    \label{fig:chap3.fig3}
\end{figure}

Figure \ref{fig:chap3.fig3} shows the number of words and documents by topic for the US newspapers.  LDA and word2vec (trained with the newspapers corpus) show homogeneous distributions of words and documents on topics, suggesting that these methods can differentiate  a set of words that compose a topic. On the other hand,  with GloVe, most of the words and documents are concentrated on a few topics.

\begin{figure}[H]
    \centering
    \includegraphics[width=.8\linewidth]{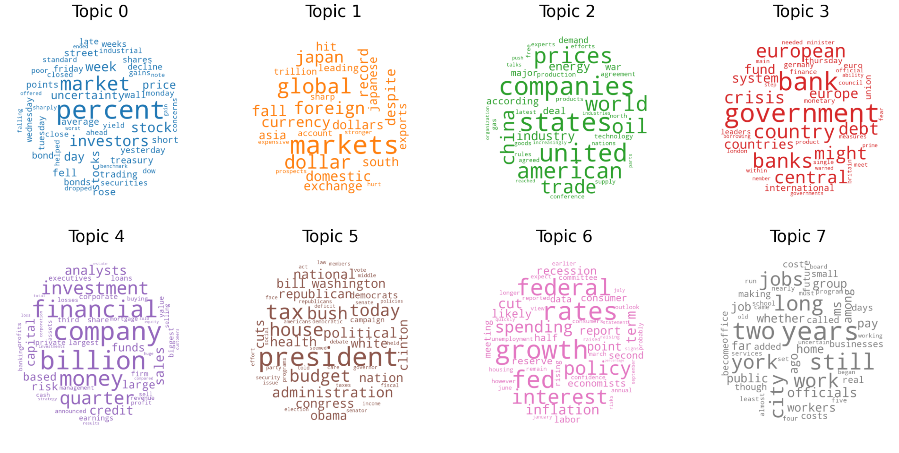}
    \caption{Word cloud by Topic (LDA US) }
    \label{fig:chap3.fig4}
\end{figure}

\begin{figure}[H]
  \centering
  \includegraphics[width=.8\linewidth]{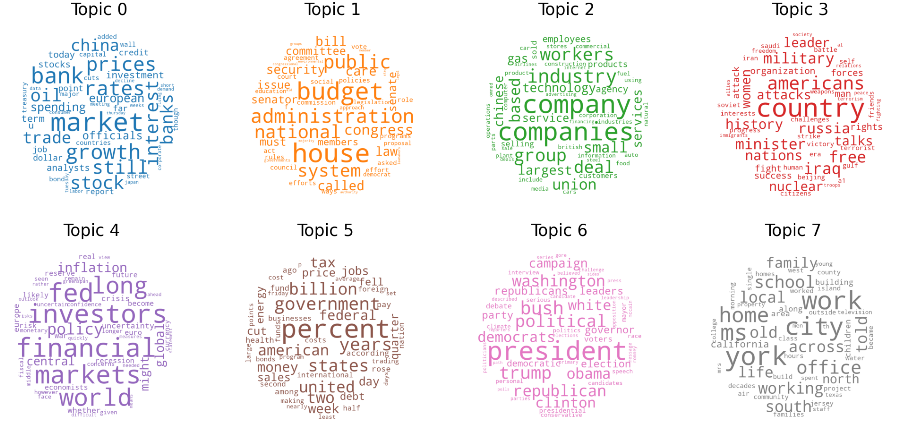}
    \caption{Word cloud by Topic (word2vec US) }
  \label{fig:chap3.fig42}
\end{figure}

\begin{figure}[H]
  \centering
  \includegraphics[width=.8\linewidth]{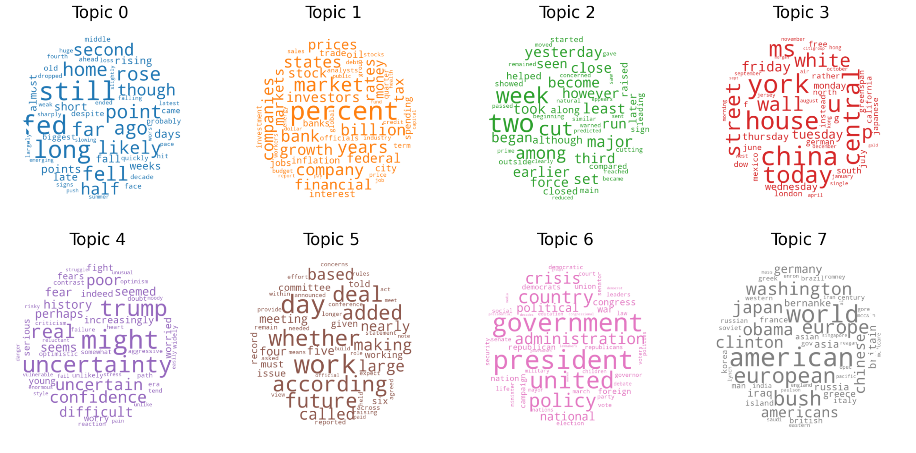}
\caption{Word cloud by Topic (word2vec-pretrained US)}  \label{fig:chap3.fig5}
\end{figure}

\begin{figure}[H]
  \centering
  \includegraphics[width=.8\linewidth]{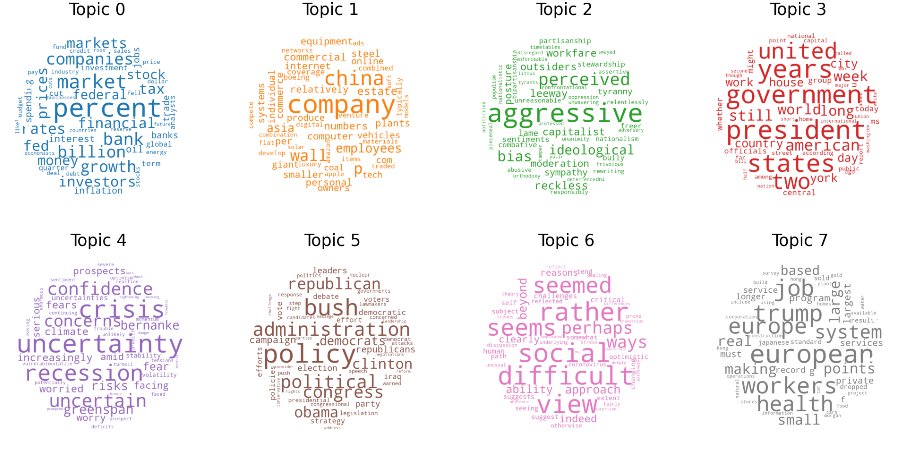}
\caption{Word cloud by Topic (GloVe US)}  \label{fig:chap3.fig6}
\end{figure}

The words assigned to each topic can be visualized in Figures \ref{fig:chap3.fig4} through \ref{fig:chap3.fig6}.  The first thing we notice is that the topics obtained with LDA are similar to those from word2vec. In contrast, by using word2vec-pretrained and GloVe,  some of LDA's topics are recovered with just a single one. Moreover, knowing the words that compose each topic, we can assign an arbitrary name, but for the purpose of this work, we named the topics according to Table \ref{tab:tabla1}, and present them in Subsection \ref{EPU subcategories}.

LDA algorithm assumes topics composed of words and documents composed of topics. The proposed approach, as mentioned in Subsection \ref{sec:methodology}, consider distance in the embedding space and allows us to cluster documents around their nearest centroid, which turns out to be, a word representation as well, in our case, a topic, thus approximating LDA method in a heuristic form. 

\begin{figure}[H] 
  \begin{minipage}[b]{0.5\linewidth}
    \centering
    \subcaption{LDA}
  \centering
  \includegraphics[width=.5\linewidth]{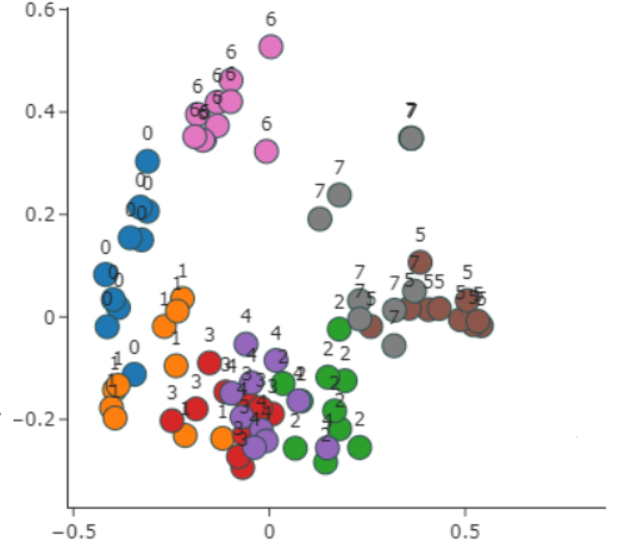}
  \centering \vspace{1ex}
  \end{minipage}
  \begin{minipage}[b]{0.5\linewidth}
    \centering
    \subcaption{word2vec}
  \centering
  \includegraphics[width=.5\linewidth]{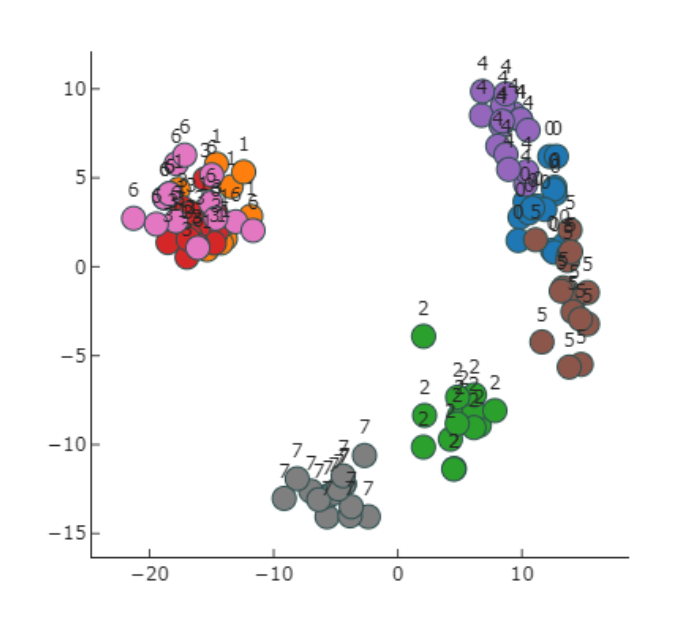}
  \centering \vspace{1ex}
  \end{minipage}
    \begin{minipage}[b]{0.5\linewidth}
    \centering
  \subcaption{word2vec-pretrained} 
  \centering
  \includegraphics[width=.5\linewidth]{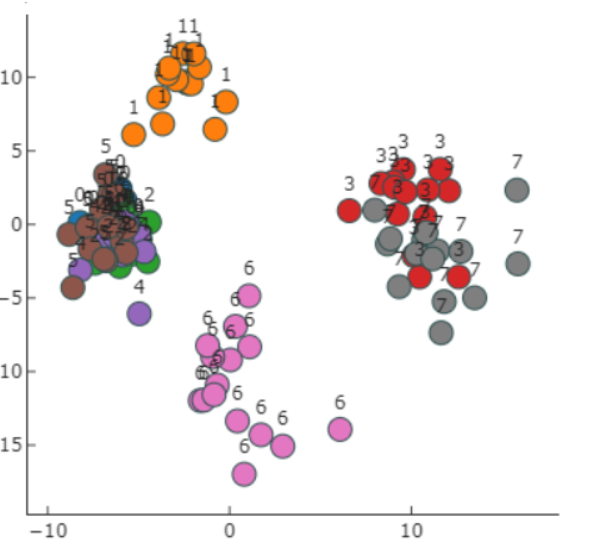}
    \vspace{1ex}
  \end{minipage}
    \begin{minipage}[b]{0.5\linewidth}
    \centering
    \subcaption{GloVe}
  \centering
  \includegraphics[width=.5\linewidth]{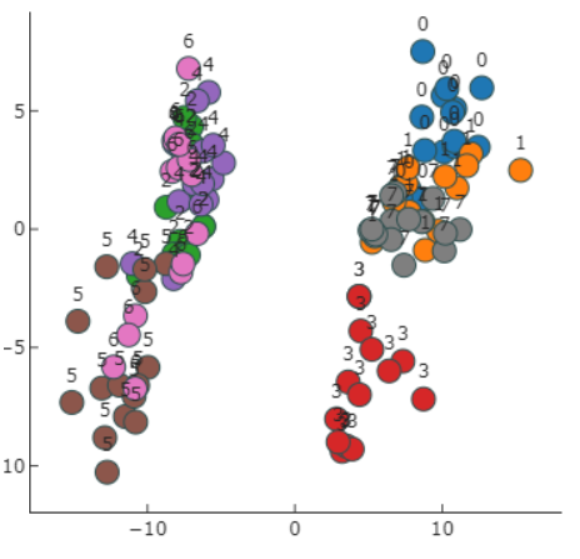}
    \vspace{1ex}
\end{minipage}

\caption{15 most representative news documents by topic (US). We define the most representative documents as those with the highest probability to belong to a a specific topic.}
    \label{fig:chap3.fig10}
\end{figure}

Now, we analyze the structure of the topic assignation of the documents by using principal component analysis (PCA) on the embedding space. For a more useful visualization, we use the most representative news documents, i.e., those with the highest probability of belonging to a certain topic.  In LDA, we only use the probability distribution of documents into topics, and in our proposed method, we use equations (\ref{eq:fuzzyK}) and (\ref{eq:fuzzyK-p}) to estimate the probability distribution based on the distance to the nearest topic (centroid of a cluster) given by the word embeddings during the learning process. Figure \ref{fig:chap3.fig10} shows the 15 most representative news items by topic. The document assignment with our approach and word2vec embeddings (\ref{fig:chap3.fig10}b and \ref{fig:chap3.fig10}c), allows us to identify interesting patterns regarding the topic structure in the US news. We can identify at least 4 clusters of news belonging to different topics, some of them are mixed, but according to the definition given in Table \ref{tab:US EPU subcategories}, it can be seen that those topics are very related. For instance, news with topics 1, 6 and 3 which are clustered together in Figure \ref{fig:chap3.fig10}b are related to regulation, fiscal policy and national security, and news with topics 0, 2, 4 and 5 clustered together in Figure \ref{fig:chap3.fig10}c, are related to financial regulation, currency crisis and uncertainty. The results obtained with LDA (\ref{fig:chap3.fig10}a) and GloVe  embeddings (\ref{fig:chap3.fig10}d) are not so informative as the previous ones. 

\subsubsection{Mexico corpus}
\label{sec:topic_mxcorpus}

\begin{figure}[H] 
    \begin{minipage}[b]{0.5\linewidth}
    \centering
  \subcaption{LDA words} 
  \centering
  \includegraphics[width=.5\linewidth]{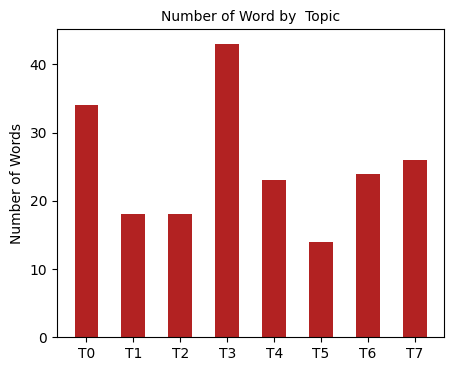}
    \vspace{1ex}
  \end{minipage}
    \begin{minipage}[b]{0.5\linewidth}
    \centering
    \subcaption{LDA news}
  \centering
  \includegraphics[width=.5\linewidth]{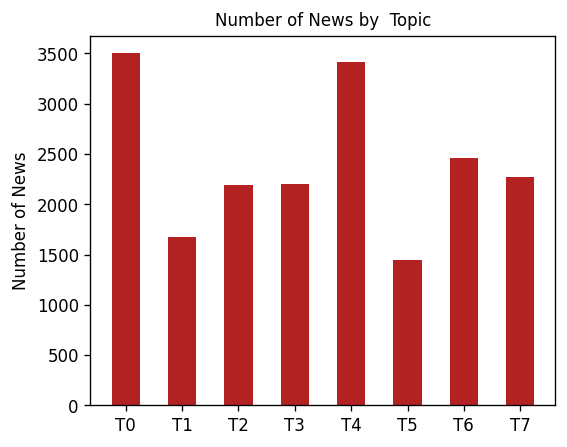}
    \vspace{1ex}
  \end{minipage}
  \begin{minipage}[b]{0.5\linewidth}
    \centering
  \subcaption{word2vec words} 
  \centering
  \includegraphics[width=.5\linewidth]{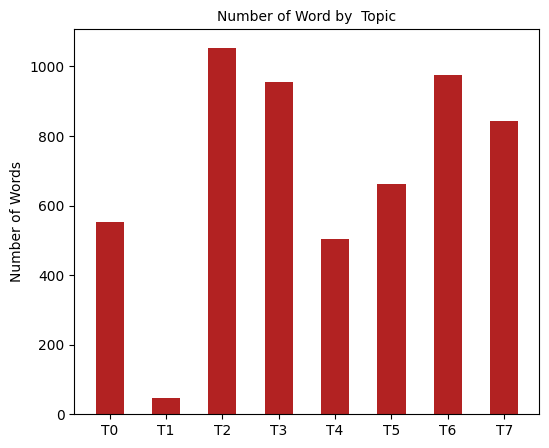}
    \vspace{1ex}
  \end{minipage}
    \begin{minipage}[b]{0.5\linewidth}
    \centering
    \subcaption{word2vec news}
  \centering
  \includegraphics[width=.5\linewidth]{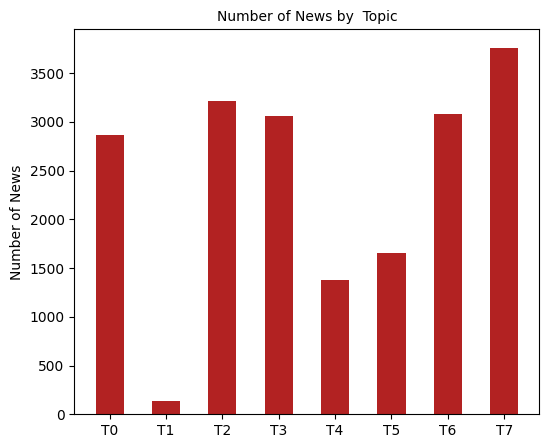}
    \vspace{1ex}
  \end{minipage}
  \begin{minipage}[b]{0.5\linewidth}
    \centering
    \subcaption{word2vec-pretrained words}
  \centering
  \includegraphics[width=.5\linewidth]{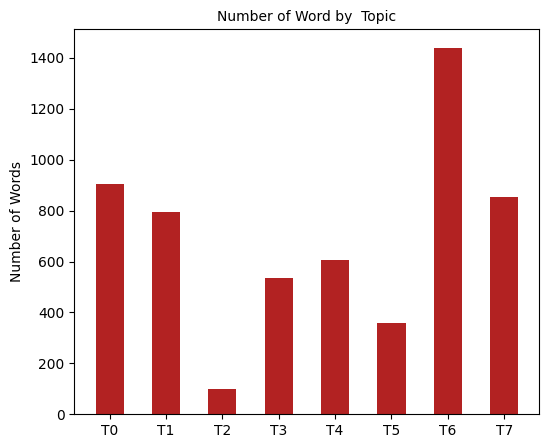}
  \centering \vspace{1ex}
  \end{minipage} 
  \begin{minipage}[b]{0.5\linewidth}
    \centering
    \subcaption{word2vec-pretrained news}
  \centering
  \includegraphics[width=.5\linewidth]{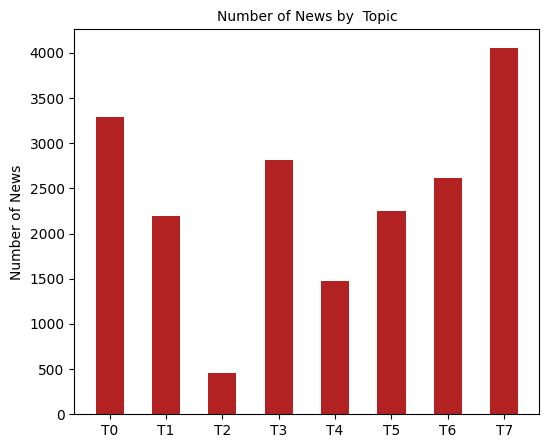}
  \centering \vspace{1ex}
  \end{minipage}
    \begin{minipage}[b]{0.5\linewidth}
    \centering
    \subcaption{GloVe words}
  \centering
  \includegraphics[width=.5\linewidth]{figs/06glove_mx_word.png}
  \centering \vspace{1ex}
  \end{minipage} 
  \begin{minipage}[b]{0.5\linewidth}
    \centering
    \subcaption{GloVe news}
  \centering
  \includegraphics[width=.5\linewidth]{figs/14glove_mx_doc.png}
  \centering \vspace{1ex}
  \end{minipage}
    \caption{Frequency of Words and Documents by Topic (Mexico)}
    \label{fig:chap3.fig11}
\end{figure}

As in the previous subsection, we compare the results with LDA and our proposed method. Figure \ref{fig:chap3.fig11} shows the frequency of words and documents assigned to each topic. LDA shows a better distribution of words and documents over the 8 topics defined previously. On the other hand, our proposal with word2vec with and without transfer learning, seems to identify less topics, as can be seen in Figures \ref{fig:chap3.fig11}(c) to \ref{fig:chap3.fig11}(f), where topics 1 and 2, which refers to inequality, fiscal policy and government, according to Table \ref{tab:MX EPU subcategories}, seems to be mixed with other topics, because there is only a few words and documents assigned to them. Interestingly, we show in Section \ref{Complement index} that it is not necessary to recover all 8 topics to reproduce the EPU index. The results with GloVe embeddings are even more extreme, because according to this model, the topic structure can be expressed only in a few of them, as can be seen in \ref{fig:chap3.fig11}(g) and \ref{fig:chap3.fig11}(h).

\begin{figure}[H]
  \centering
  \includegraphics[width=.8\linewidth]{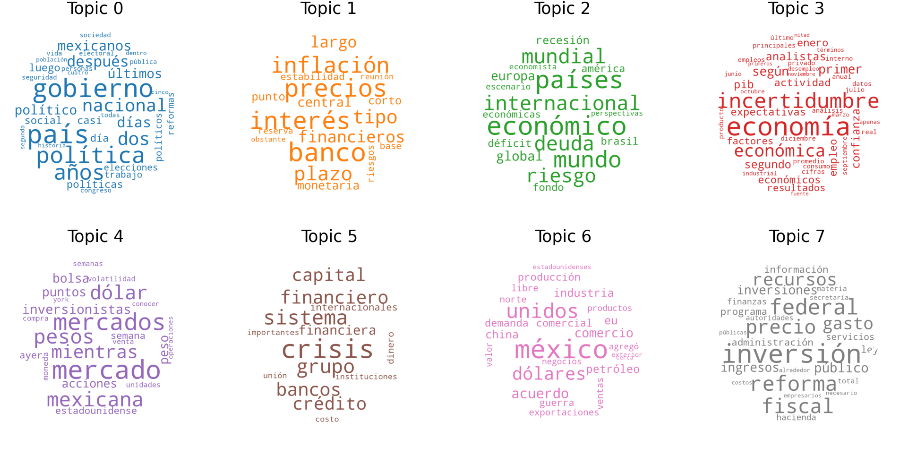}
  \caption{Word Clouds Words by Topic (LDA Mexico) }
  \label{fig:chap3.fig12}
\end{figure}

\begin{figure}[H]
  \centering
  \includegraphics[width=.8\linewidth]{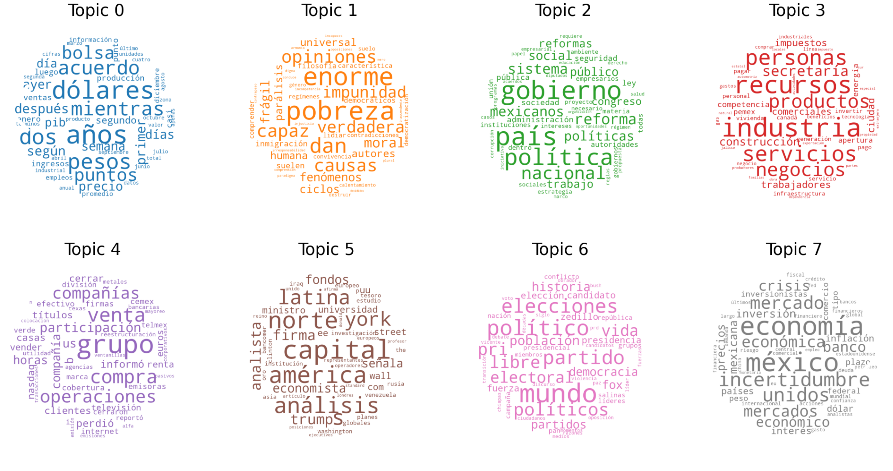}
  \caption{Word Clouds Words by Topic (word2vec Mexico) }
  \label{fig:chap3.fig12a}
\end{figure}

\begin{figure}[H]
  \centering
  \includegraphics[width=.8\linewidth]{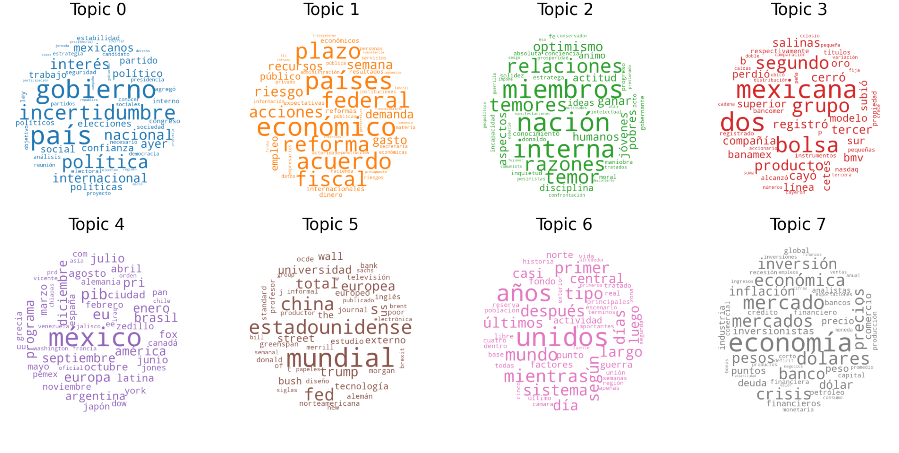}
  \caption{Word Clouds Words by Topic (word2vec-pretrained Mexico)}
  \label{fig:chap3.fig13}
\end{figure}

\begin{figure}[H]
  \centering
  \includegraphics[width=.8\linewidth]{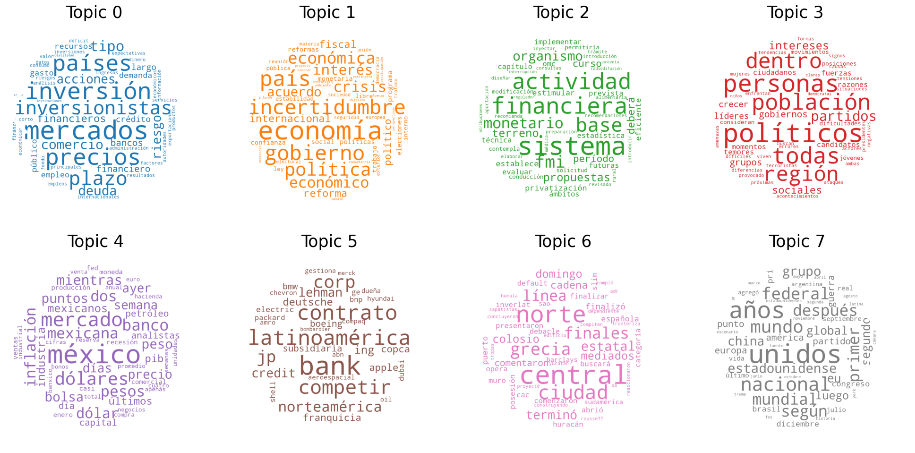}
  \caption{Word Clouds Words by Topic (GloVe Mexico)}
  \label{fig:chap3.fig14}
\end{figure}

Figures \ref{fig:chap3.fig12} through \ref{fig:chap3.fig14} shows the word clouds by topic for Mexico news. In general, LDA and the proposed method with different word embeddings found similar topics, and many words assigned to the topics by the LDA probability distribution are in one of the clusters obtained with the proposed method.
 
\begin{figure}[H] 
  \begin{minipage}[b]{0.5\linewidth}
    \centering
    \subcaption{LDA}
  \centering
  \includegraphics[width=.5\linewidth]{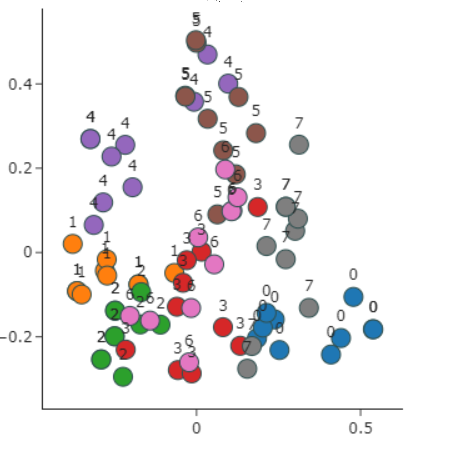}
  \centering \vspace{1ex}
  \end{minipage}
  \begin{minipage}[b]{0.5\linewidth}
    \centering
    \subcaption{word2vec}
  \centering
  \includegraphics[width=.5\linewidth]{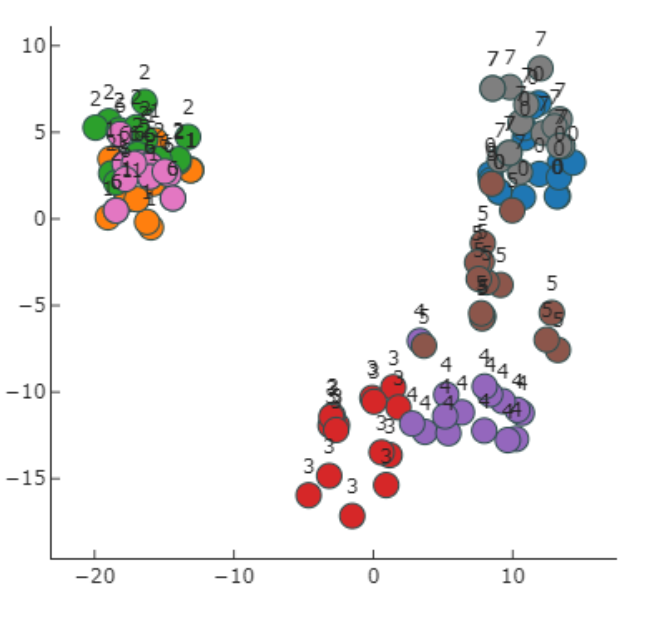}
  \centering \vspace{1ex}
  \end{minipage}
    \begin{minipage}[b]{0.5\linewidth}
    \centering
  \subcaption{word2vec-pretrained} 
  \centering
  \includegraphics[width=.65\linewidth]{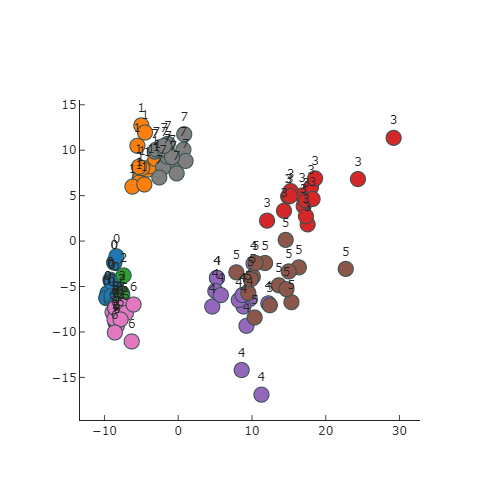}
    \vspace{1ex}
  \end{minipage}
    \begin{minipage}[b]{0.5\linewidth}
    \centering
    \subcaption{GloVe}
  \centering
  \includegraphics[width=.65\linewidth]{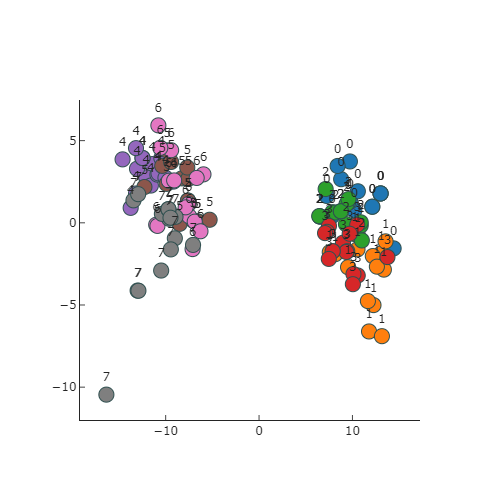}
    \vspace{1ex}
  \end{minipage}
    \caption{15 most representative news documents by topic (Mexico). We define the most representative documents as those with the highest probability to belong to a a specific topic.}

    \label{fig:chap3.fig18}
\end{figure}

Finally, we compare the structure of topic assignation by visualizing the projections in the first two components of PCA. As in the US news, we select the 15 most representative news documents by topic, which can be seen in Figure \ref{fig:chap3.fig18}. We can see that LDA's representation overlaps  the documents over the topics, meanwhile with the proposed model and word2vec embeddings (Figure \ref{fig:chap3.fig18}(b) and (c)) we can identify interesting patterns about the topics and the related news documents, and even the topics mixed in some clusters are very related according to the definition of Table \ref{tab:MX EPU subcategories}, for instance, topic 1 and 7 in Figure \ref{fig:chap3.fig18}(b) which refers to fiscal policy.

\subsection{Time comparison}
\label{sec:computing-time}
\begin{figure}[H] 
  \begin{minipage}[b]{0.5\linewidth}
    \centering
  \subcaption{LDA} 
  \centering
  \includegraphics[width=.8\linewidth]{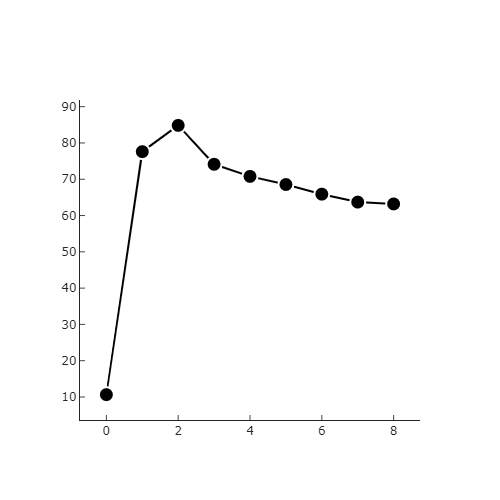}
    \vspace{1ex}
  \end{minipage}
    \begin{minipage}[b]{0.5\linewidth}
    \centering
    \subcaption{word2vec}
  \centering
  \includegraphics[width=.8\linewidth]{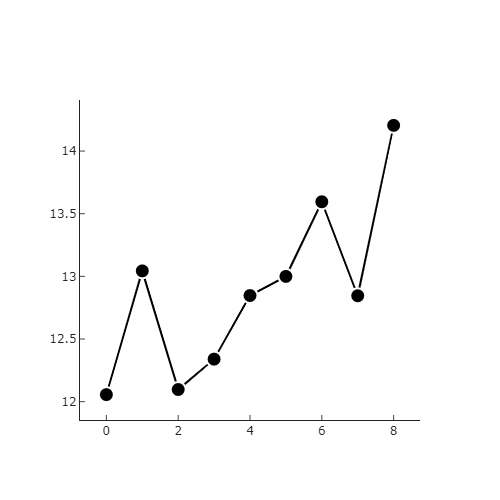}
    \vspace{1ex}
  \end{minipage}
  \caption{Computing time. The y-axis is in seconds, and the x-axis is the number of topics $k$.}
      \label{fig:chap3.4}
\end{figure}

One advantage of the proposed approach is the time required to calculate the topic structure. For comparison purposes, we present in Figure \ref{fig:chap3.4} the computing time in seconds (y-axis) as a function of the number of topics $k$ (x-axis) for LDA and our proposal based on fuzzy $k-$means with word2vec embeddings. For LDA, the mean time is around 70 seconds, however, if we use our approach, once the embedding is trained, the time required for clustering in the semantic space is shorter and increases very slightly as $k$ increases. 

\subsection{Dynamic topic assignation} 

One of the main advantages of our proposal is the ability to make dynamic assignation of news documents to topics. As we mentioned in Section \ref{sec:methodology}, once we obtain a cluster (or topic) configuration based on a training corpus, we can take a news document, find its representation in the embedding space, and make the topic assignation according to the distance to the nearest centroid in the cluster configuration, which corresponds to the topic with the maximum probability (see Eq. \ref{eq:fuzzyK} and \ref{eq:fuzzyK-p}). In this work, we used cosine distance, but another disimilarity measure can be considered, such as Euclidean distance. 

One limitation of using pretrained word embedding such as word2vec or GloVe, is that only the words in the training vocabulary are considered, so, if there is an out of vocabulary (OOV) word in an new document, it will not be possible to obtain an embedding. There are some options to deal with this problem. The easiest one is to enlarge the training corpus with the new documents and re-estimate the topics configuration by using transfer learning according to the proposed methodology in Section \ref{sec:methodology}. Certainly, this approach is the same if we use LDA to obtain the topics, where the algorithm must be applied again for topic assignation of new documents, but taking advantage of the fast training time of our proposal compared to LDA (see Section \ref{sec:computing-time}), this option is feasible. Another options to deal with OOV words, is to give a constant value (such as the average) to the embeddings, or using the embedding of the most similar word in the vocabulary according to some string distance. A more efficient approach is to use some embeddings model which can deal with OOV words, such as fastText \citep{fasttext_bojanowski-etal-2017}. In this work, we use the first approach to illustrate the dynamic topic assignation, but the implementation of other alternatives are considered as future work. 

Figure \ref{fig:chap3.fig22} presents the results for US newspapers from April 1 to April 22, 2020, by using word2vec and GloVe embeddings. News items are clustered with their nearest centroids, and the reason that documents overlap with other topics is that a single document has a probability of belonging to each topic. In this example, we use the cosine distance to make a hard assignation. 

\begin{figure}[H] 
   \begin{minipage}[b]{0.5\linewidth}
    \centering
  \subcaption{word2vec} 
  \centering
  \includegraphics[width=.6\linewidth]{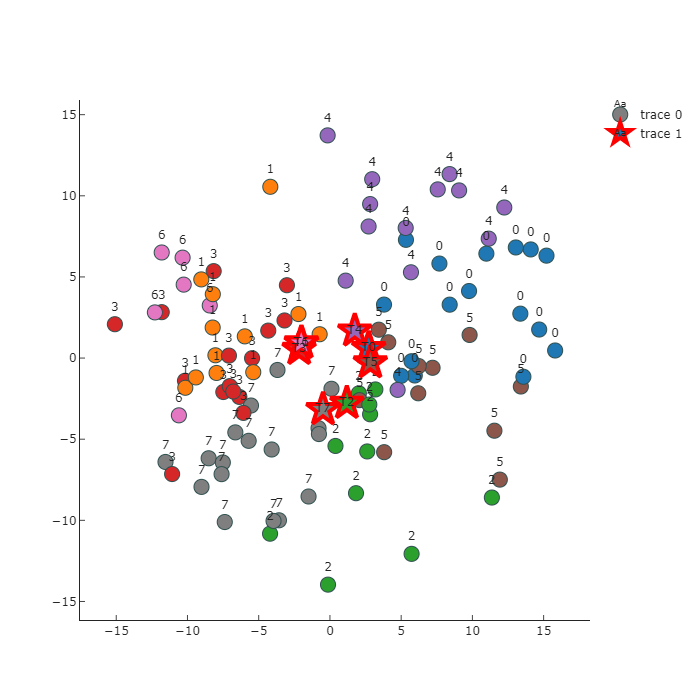}
    \vspace{1ex}
  \end{minipage}
    \begin{minipage}[b]{0.5\linewidth}
    \centering
  \subcaption{word2vec-pretrained} 
  \centering
  \includegraphics[width=.6\linewidth]{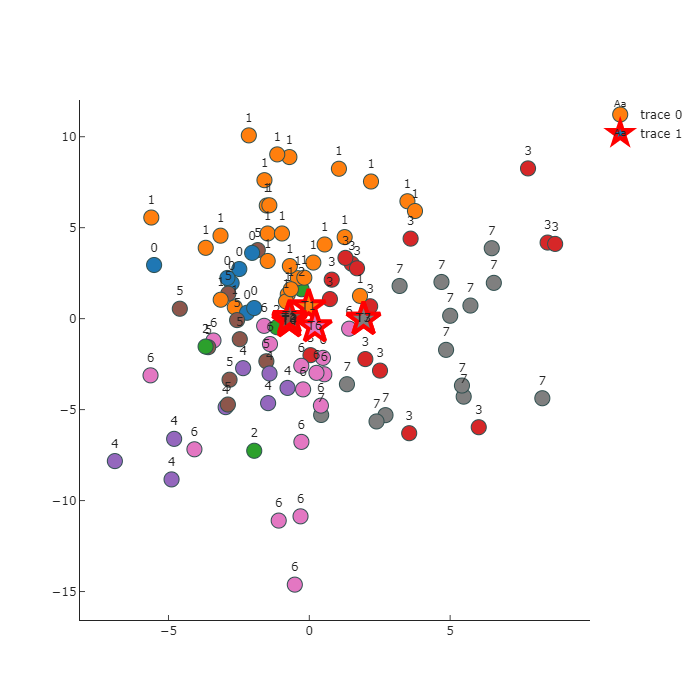}
    \vspace{1ex}
  \end{minipage}
       \centerline{\begin{minipage}[b]{0.5\linewidth}
    \centering
    \subcaption{GloVe}
  \centering
  \includegraphics[width=.6\linewidth]{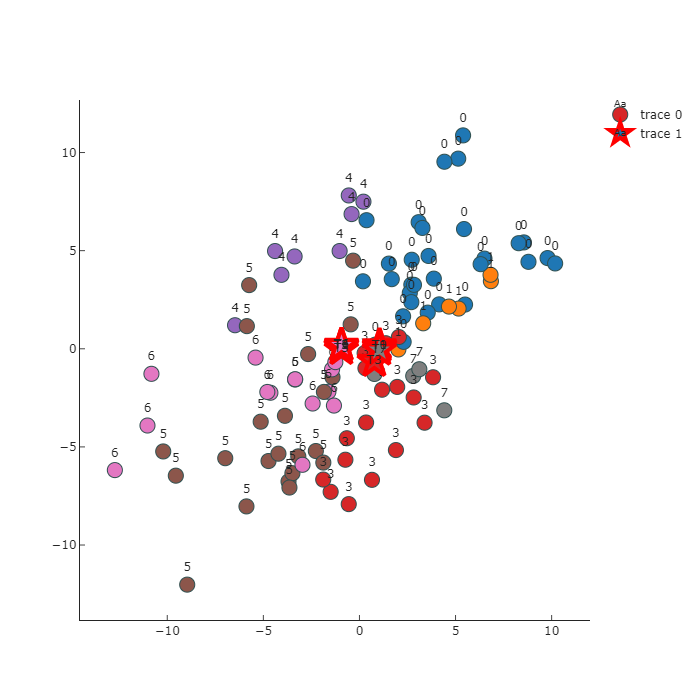}
    \vspace{1ex}
  \end{minipage}}
    \caption{Dynamic topic assignation (US news from April 1 to April 22 2020). We show the first two components of PCA. The stars are the centroids of the previously trained word representation.}
    \label{fig:chap3.fig21}
\end{figure}

Figure \ref{fig:chap3.fig22} presents the dynamic topic assignation for Mexico news. The news comes from the same period of time as that of the US news, and for the assignations, we use the cosine distance as well. We see similar results to those for the US: news items are classified and some of them overlap. However, as we previously noted, news items are related to many topics, and in this case, we classify them with the most relevant topic.

\begin{figure}[H] 
 \begin{minipage}[b]{0.5\linewidth}
    \centering
  \subcaption{word2vec} 
  \centering
  \includegraphics[width=.6\linewidth]{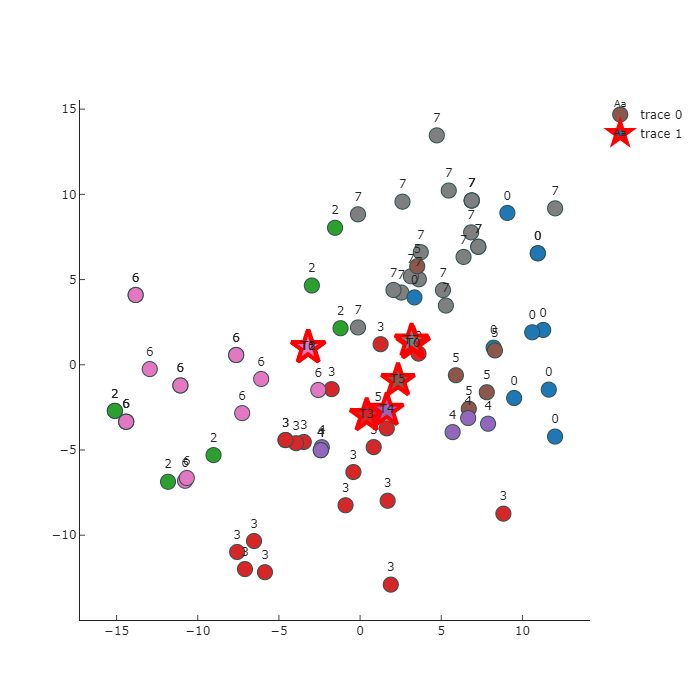}
    \vspace{1ex}
  \end{minipage}
    \begin{minipage}[b]{0.5\linewidth}
    \centering
  \subcaption{word2vec-pretrained} 
  \centering
  \includegraphics[width=.6\linewidth]{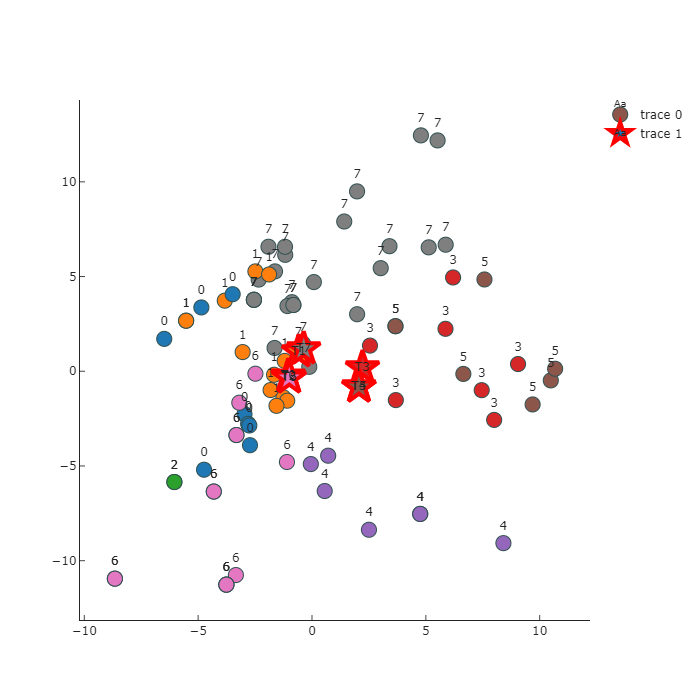}
    \vspace{1ex}
  \end{minipage}
   \centerline{\begin{minipage}[b]{0.5\linewidth}
    \centering
    \subcaption{GloVe}
  \centering
  \includegraphics[width=.6\linewidth]{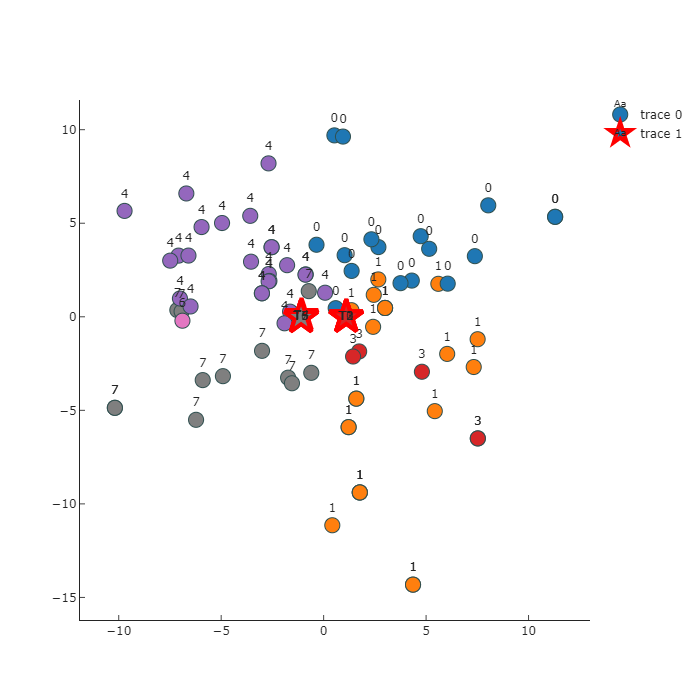}
    \vspace{1ex}
  \end{minipage}}
    \caption{Dynamic topic assignation (Mexico from April 1 to April 22, 2020). We show the first two components of PCA. The stars are the centroids of the previously trained word representation.}
  
    \label{fig:chap3.fig22}
\end{figure}

\subsection{EPU subcategories} \label{EPU subcategories}

The previous analysis focused on comparing LDA and our proposed method, specifically for the estimation of the news topic distribution. As we mentioned previously in Section \ref{sec:methodology}, Baker's EPU index is the combination of the dynamics of some subcategories, represented as time series of topics presented in digital news. In this subsection, we analyze our results respecting to the EPU subcategories (step 4 of the algorithm presented in Section \ref{sec:methodology}) for US and Mexico digital newspapers.

\subsubsection{US corpus}

Table \ref{tab:US EPU subcategories} shows the US EPU subcategories. The topic names, as we mentioned before, were selected following those of Baker and Azqueta-Gavaldón and looking for the words that compose the topic (Table \ref{tab:tabla1}). In some cases, the topics obtained with our proposal coincide with Baker's, and with some words representations, the topics are composed of many of Baker's. However, an interesting finding is that it is not necessary to obtain the same topics in order to reproduce the EPU index. Actually, we can generate more informative topics. According to Table \ref{tab:tabla1} and the word clouds of Section \ref{sec:topic_uscorpus}, the most similar topics to those of LDA were those estimated with word2vec embeddings (without pretrained weights). 

\begin{table}[H]
\tiny
\resizebox{\textwidth}{!}
{ \begin{tabular}{llllll}
    \toprule
       Topics  & LDA &word2vec   & word2vec-pretrained & GloVe   \\
   \midrule
   Topic0 & Financial regulation & Trade Policy & Financial regulation & Financial regulation \\
   &(stock market& &(stock market)&and Monetary- Fiscal Policy\\
    & and , financial investment) &  &  & (stock market, financial investment)\\
    \hline
        Topic1 & Exchange rate policy & Regulation and Fiscal Policy & Financial regulation & Trade Policy  debt\\
        &&(taxes, government spending)&and Monetary- Fiscal-Trade Policy&and Sovereign\\
        
      & & &  (financial investment, energy) &\\
       \hline
        Topic2 & Trade Policy & Sovereign debt & Currency crisis & Ideology\\\
          &&&&\\
          & & &  &\\\\
           \hline
       Topic3 & Currency crisis& National Security & Currency crisis &  Regulation\\
       &&(conflict, Iraq, Russia)&&\\
         & & &  &\\
          \hline
       Topic4  & Sovereign debt &Monetary Policy& Uncertainty& Current crisis and Uncertainty\\
       &&&&\\
          & && & \\
           \hline
        Topic5 & Regulation and Fiscal Policy &Regulation &Financial regulation   & Regulation \\
         &  (taxes, government spending) &(stock market&  &  (law)\\
        &&and  financial investment)&&\\
          \hline
       Topic6 & Monetary Policy &Regulation& Regulation   &Wellness\\ 
         &(laws)&(law)&&(social)\\
         & & & &\\ 
          \hline
      Topic7 & Wellness &Wellness& National
Security &Wellness\\   

 & (education , health, programs)&(education, health, programs) &  (conflict, Russia)&\\   
  &&&&\\
\bottomrule
\end{tabular}}
\caption{US EPU subcategories}
\label{tab:US EPU subcategories}
\end{table}

The name of the topic could be assigned non-arbitrarily. In our approach, we can use the embedding space for the words on each cluster, or just the $N$ most frequent. In any case, we take their corresponding embedding vectors obtained with the trained model to compute the centroid of those embeddings and select the nearest neighbour embedding from the vocabulary based on the cosine distance. The name of the topic will be the word which corresponds to the this embedding. The problem with this approach is that sometimes the name of the topic is not very informative, unless we restrict the words from the vocabulary we will use for this purpose. 

\begin{figure}[H]
  \centering
  \includegraphics[width=.8\linewidth]{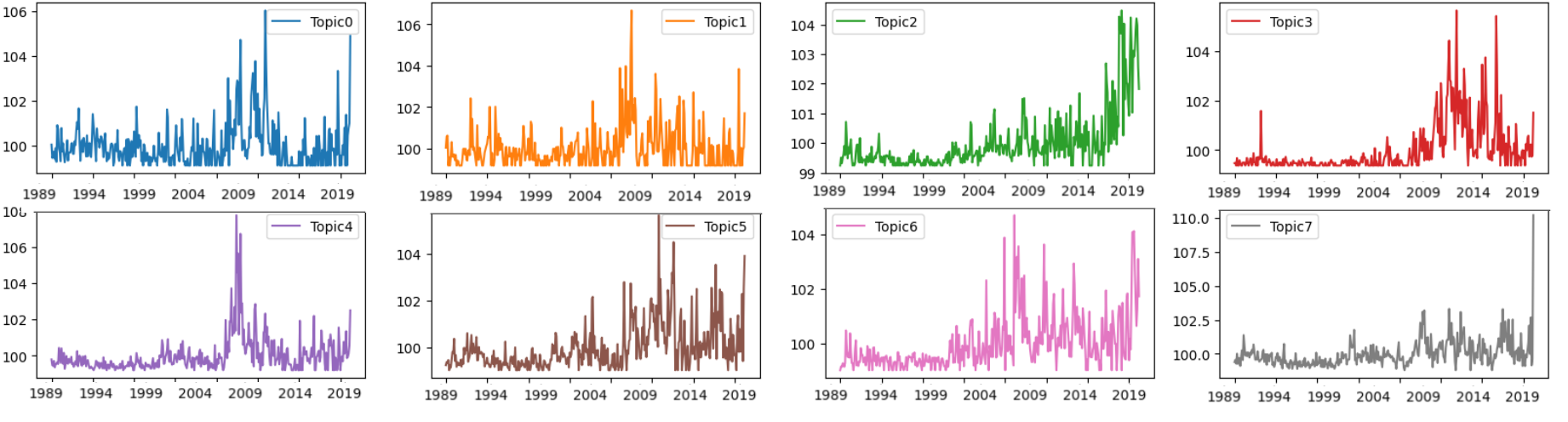}
    \caption{US EPU subcategories. LDA}  
  \label{fig:chap3.fig23}
\end{figure}

\begin{figure}[H]
  \centering
  \includegraphics[width=.8\linewidth]{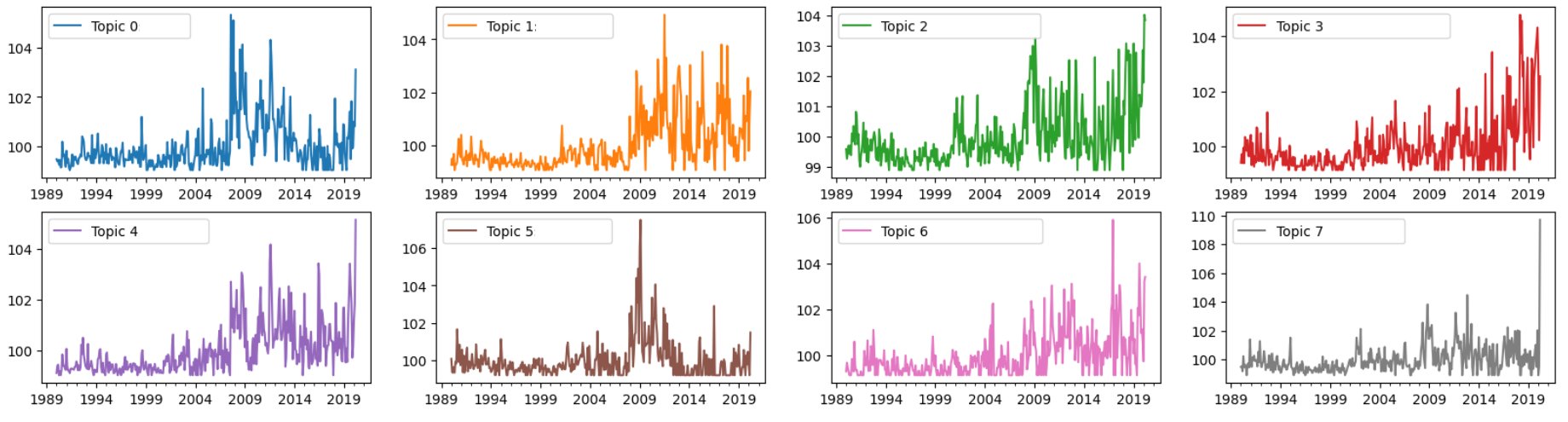}
\caption{US EPU subcategories. word2vec}
  \label{fig:chap3.fig23extra}
\end{figure}

\begin{figure}[H]
  \centering
  \includegraphics[width=.8\linewidth]{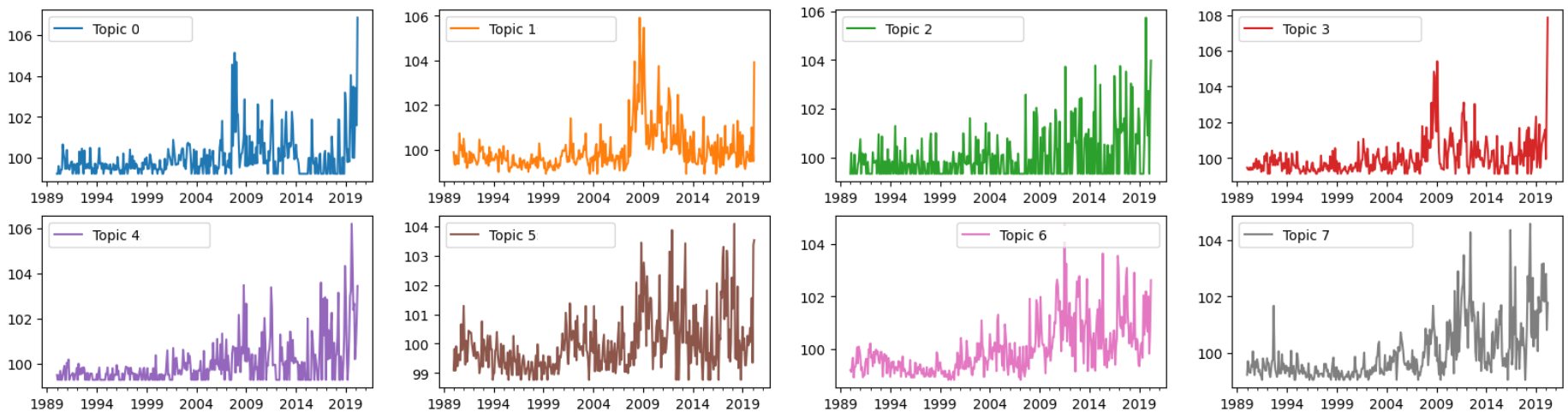}
\caption{US EPU subcategories. word2vec-pretrained}  
  \label{fig:chap3.fig24}
\end{figure}

\begin{figure}[H]
  \centering
  \includegraphics[width=.8\linewidth]{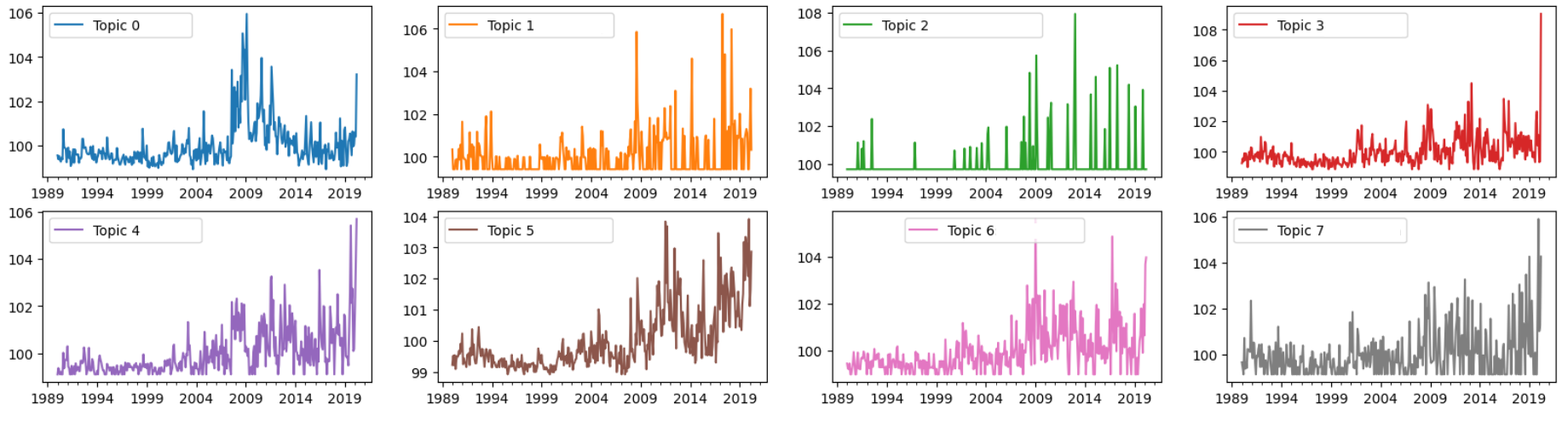}
\caption{US EPU subcategories. GloVe}
  \label{fig:chap3.fig25}
\end{figure}

Figures \ref{fig:chap3.fig23} through \ref{fig:chap3.fig25} shows the US EPU subcategories from January 1990 to March 2020. We show time series of the topics we obtained by using LDA and the proposed method. One of the effects of assigning few documents to one topic can be seen in the results obtained with GloVe embeddings in Figure \ref{fig:chap3.fig25}. Topic 2 (ideology) has few news items, suggesting that it is not an informative topic by itself. Later in Section \ref{Complement index}, we present a methodology to create new subcategory indexes by aggregating the time series of topics based on a distance criteria in the embedding space.

\subsubsection{Mexico corpus}

In this case, Azqueta's methodology is not applied to Mexico, only Baker created an EPU index for Mexico. In this sense, a significant contribution to the literature is the automation of Mexico index. 
As we did before for US, we tried to recover the topics from Table \ref{tab:tabla1}, and the results are shown in Table  \ref{tab:MX EPU subcategories}, where we describe the subcategory names by method.  Many topics are the same as those of the US: fiscal, monetary and trade policy, crisis and sovereign debt. Furthermore, the Mexican news uncertainty is related to other topics and  the economic activity in terms of national production.  

\begin{table}[H]
\footnotesize
\centering
\begin{tabular}{llllll}
\toprule
   Topics  & LDA &word2vec   & word2vec-pretrained & GloVe   \\
   \midrule
   Topic0 &Politic&Financial
regulation&Politic &Fiscal Policy\\
   &&&&\\
   &&&&\\
   \hline
   Topic1 &Monetary Policy&Inequality&Fiscal Policy&Economic activity\\
    &&(stock market)&&\\
    &&&&\\
    \hline
    Topic2 &Sovereign debt &Goverment Programs&Inequality&Sovereign debt
 \\
    &&&&and currency crisis\\
    &&&&(financial crisis)\\
    \hline
    Topic3 &Economy activity&Economy activity&Financial
regulation&Politic\\
    &&&(stock market)&\\
    &&&&\\
    \hline
    Topic4  &Financial
regulation&Trade Policy&Global economy&Monetary Policy\\
    &(stock market)&&&\\
    &&&& \\
    \hline
    Topic5 &Sovereign debt&Sovereign debt&Trade Policy&Financial
regulation \\
    &and currency crisis&and currency crisis&&\\
    &(financial crisis)&(financial crisis)&&(stock market)\\
    \hline
    Topic6 &Trade Policy&Politic&Global economy&Global economy\\ 
    &&&&\\
    &&&&\\ 
    \hline
    Topic7 &Fiscal Policy&Fiscal-Monetary Policy&Fiscal-Monetary Policy &Trade Policy\\   
    &&and Sovereign debt&and Sovereign debt&\\   
    &&&&\\
\bottomrule
\end{tabular}
  \caption{Mexico EPU subcategories}
\label{tab:MX EPU subcategories}
\end{table}

\begin{figure}[H]
  \centering
  \includegraphics[width=.8\linewidth]{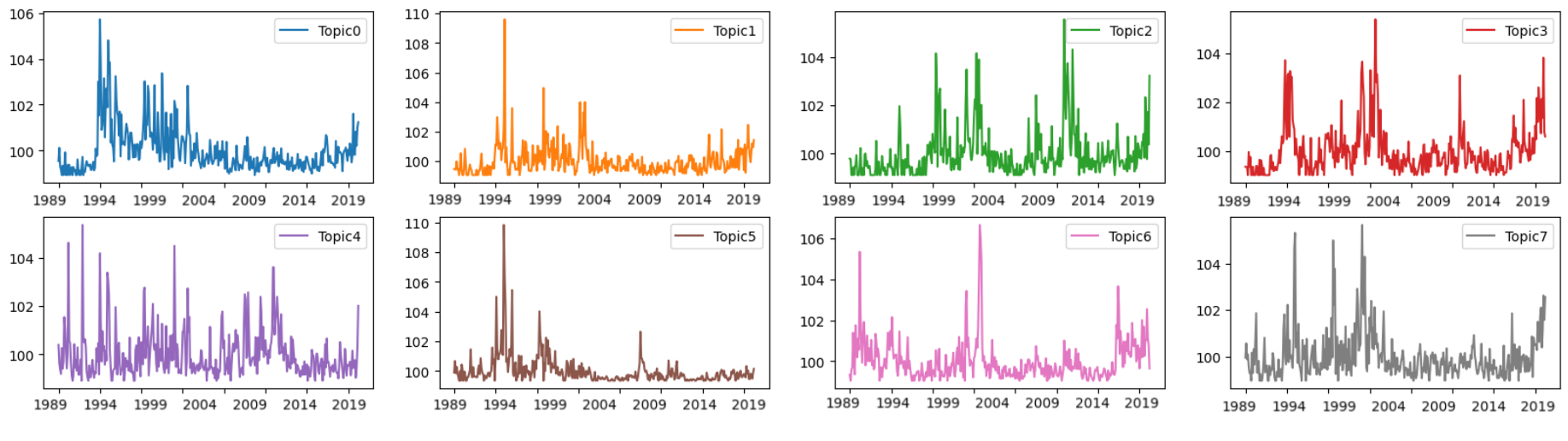}
\caption{LDA}  
  \label{fig:chap3.fig26}
\end{figure}

\begin{figure}[H]
  \centering
  \includegraphics[width=.8\linewidth]{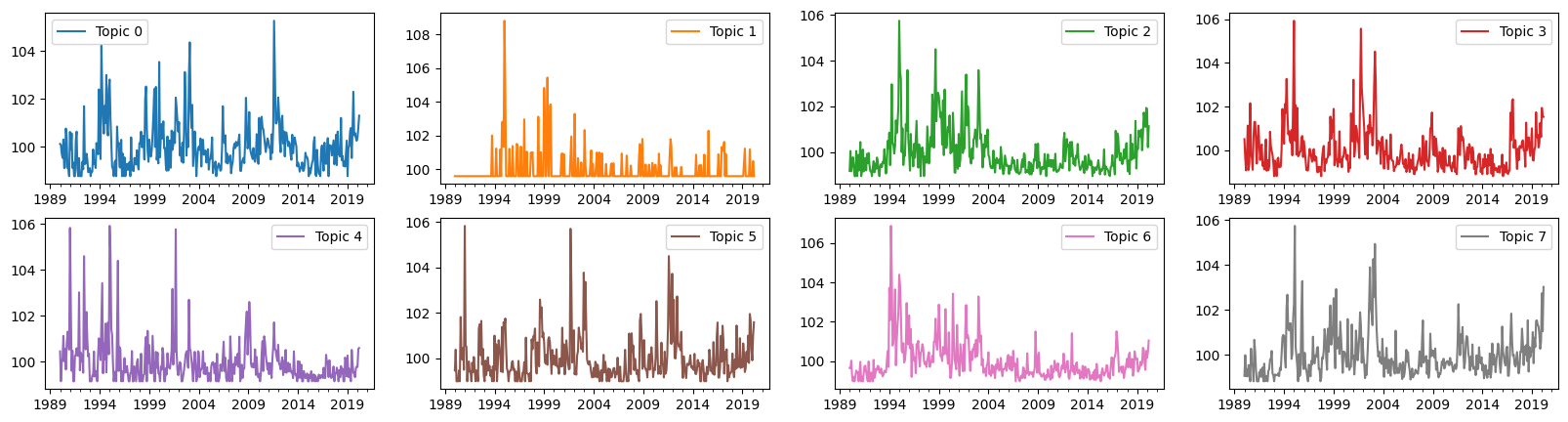}
\caption{word2vec}
\label{fig:chap3.fig26-2}
\end{figure}

\begin{figure}[H]
  \centering
  \includegraphics[width=.8\linewidth]{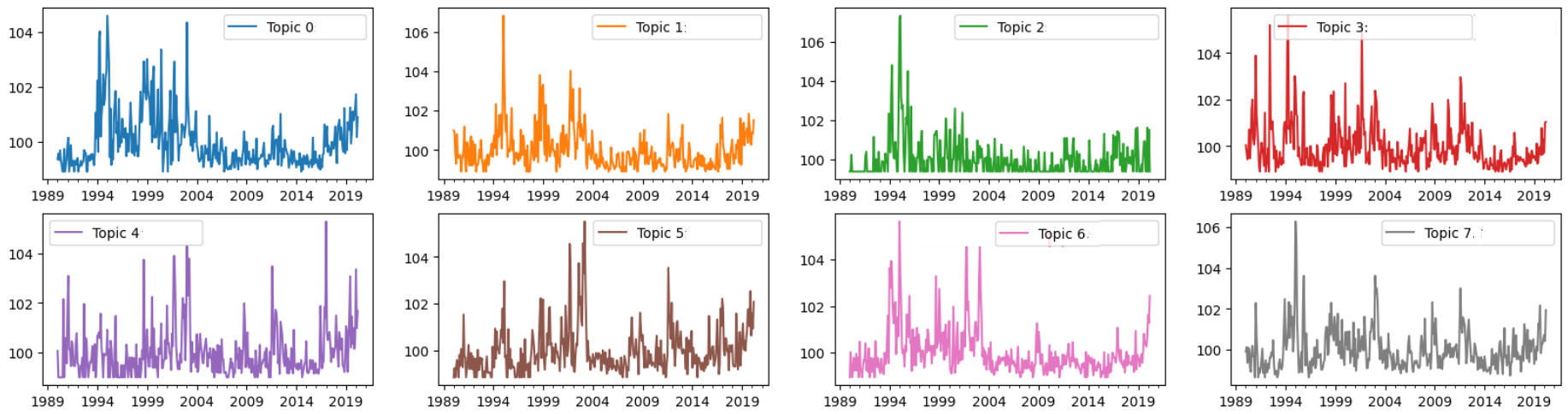}
\caption{word2vec-pretrained}
  \label{fig:chap3.fig27}
\end{figure}

\begin{figure}[H]
  \centering
  \includegraphics[width=.8\linewidth]{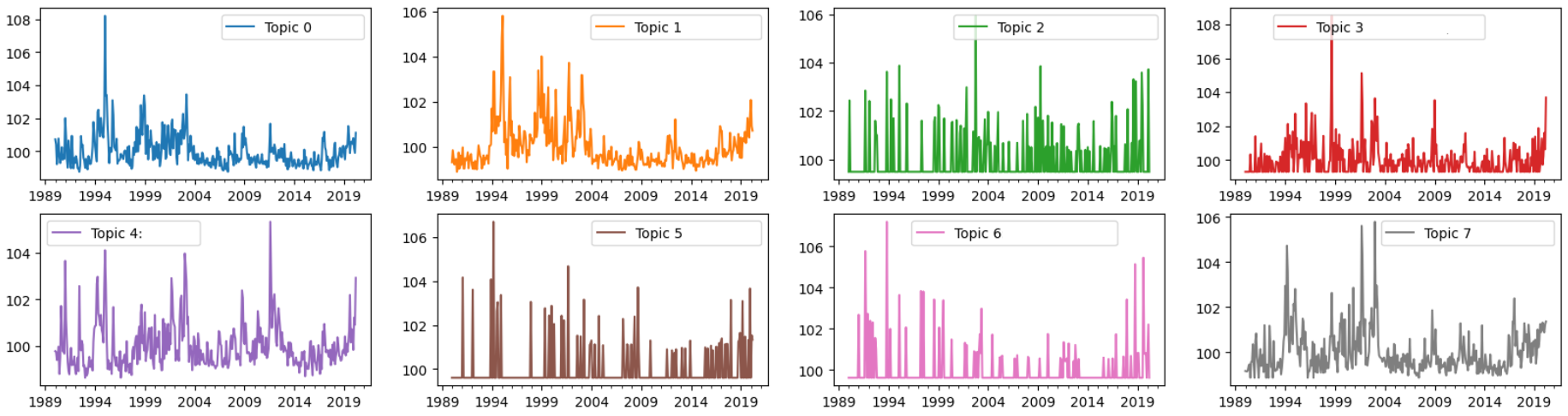}
\caption{GloVe}
\label{fig:chap3.fig28}
\end{figure}

Figures \ref{fig:chap3.fig26} through \ref{fig:chap3.fig28}  present the Mexico EPU subcategories by method. As in the US case, topics with a small amount of news items are not very informative, as is the case of the topics obtained with GloVe embeddings in Figure \ref{fig:chap3.fig28}. Also, in Section \ref{Complement index} we will show how we can group subcategories to generate a more informative index. 

\subsection{EPU index} 

\begin{figure}[H] 
    \begin{minipage}[b]{0.33\linewidth}
    \centering
    \includegraphics[width=.9\linewidth]{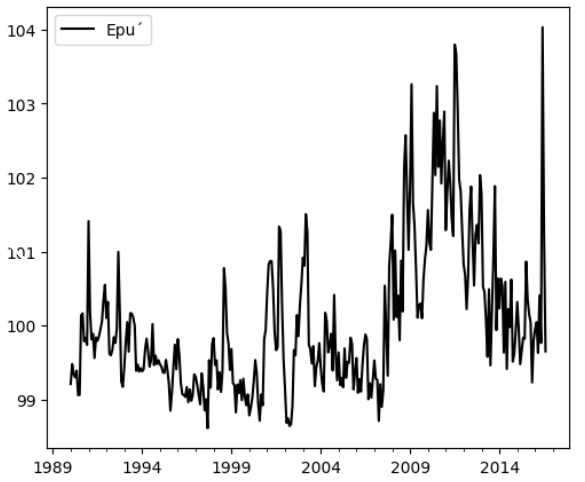}
    \subcaption{EPU-Baker} 
    \end{minipage}
    \begin{minipage}[b]{0.33\linewidth}
    \centering
    \includegraphics[width=.9\linewidth]{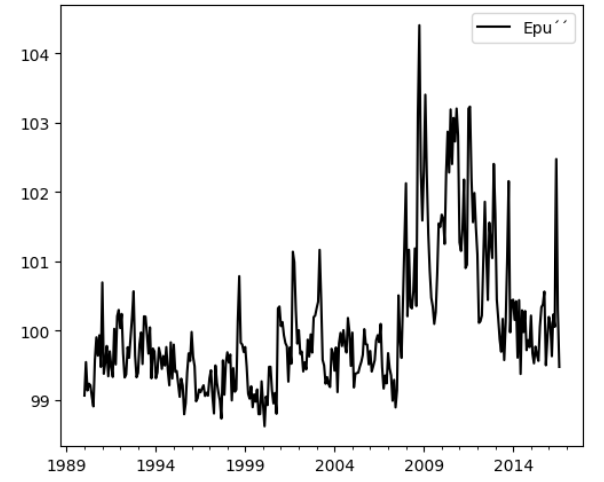}
    \subcaption{EPU-Azqueta}
    \end{minipage}
    \begin{minipage}[b]{0.33\linewidth}
    \centering
    \includegraphics[width=.9\linewidth]{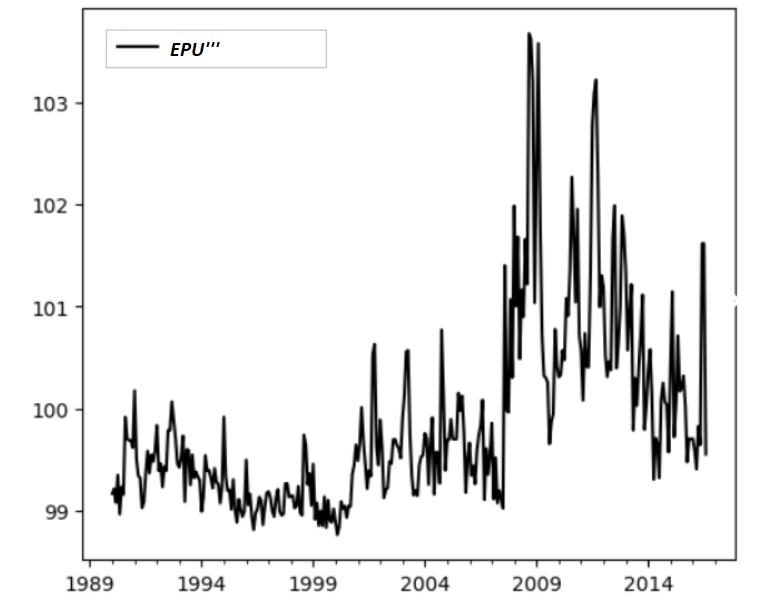}
    \subcaption{EPU-word2vec} 
    \end{minipage}
\caption{EPU Index for US, from January 1989 to October 2017}
\label{fig:chap3.fig29}
\end{figure}

In this subsection, we introduce the EPU index and its components. In order to compare our approach with Baker's and Azqueta's, we will cover the period from January 1990 to October  2017, because Baker's index is available only for this period of time. Figure \ref{fig:chap3.fig29} shows the EPU index for US. We refer to Baker's index as EPU-Baker, Azqueta's index as EPU-Azqueta, and our proposed index as EPU-word2vec, where we use word2vec embeddings without pretrained weights following the methodology we presented in Section \ref{sec:methodology}. We decided to use this word embeddings because it shows consistently better results and there is no significant differences by using word2vec with transfer learning. The  EPU-word2vec index recovers the tendency of the other approaches where we can identify some relevant events related to uncertainty: the terrorist attacks in 9/11 2001, the Gulf War II in 2003, the global crisis in 2008, and Trump's election in 2016. The difference among indexes is the level of volatility they capture. The EPU-Baker is more volatile than  EPU-Azqueta and EPU-word2vec, as can be seen in the peak related to Trump's election. This behaviour is more notable in EPU-Baker since  EPU-Azqueta and EPU-word2vec reduce the uncertainty.

In order to perform quantitative comparisons between the indexes, we use the correlation and dynamic time warping (DTW) statistic \citep{Muller2007} to measure the similarity between the time series. Tables \ref{table:dtwUS} and \ref{table:correlationUS} presents the results for DTW and correlation, respectively. Small positive values of DTW indicate greater similarity between the indexes, and according to the results in Table \ref{table:dtwUS}, EPU-Azqueta and EPU-word2vec indexes are very similar, but with our proposal, we compute the index with a significant reduction in computational cost.   

\begin{table}[h]
\footnotesize
 \centering
 \begin{tabular}{llllll}
\toprule
       & 	EPU-Baker	&EPU-Azqueta	\\
   \midrule
EPU-Baker &\multicolumn{1}{c}{-}	&5.218 \\
EPU-Azqueta &5.218	&\multicolumn{1}{c}{-} \\
EPU-word2vec  &5.846	&6.885\\
\bottomrule
\end{tabular}
\caption{Dynamic Time Warp: US}
\label{table:dtwUS}
\end{table}    

\begin{table}[h]
\footnotesize
 \centering
 \begin{tabular}{llllllll}
    \toprule
       & 	EPU-Baker	&EPU-Azqueta	\\
   \midrule
EPU-Baker	&1.0000	&0.9374	\\
EPU-Azqueta	&0.9374	&1.0000	\\
EPU-word2vec 	&0.8406	&0.8683		\\
\bottomrule
\end{tabular}
\caption{Correlation: US}
\label{table:correlationUS}
\end{table}    

\begin{figure}[H] 
    \begin{minipage}[b]{0.5\linewidth}
    \centering
    \subcaption{EPU-Baker} 
    \centering
    \includegraphics[width=.8\linewidth]{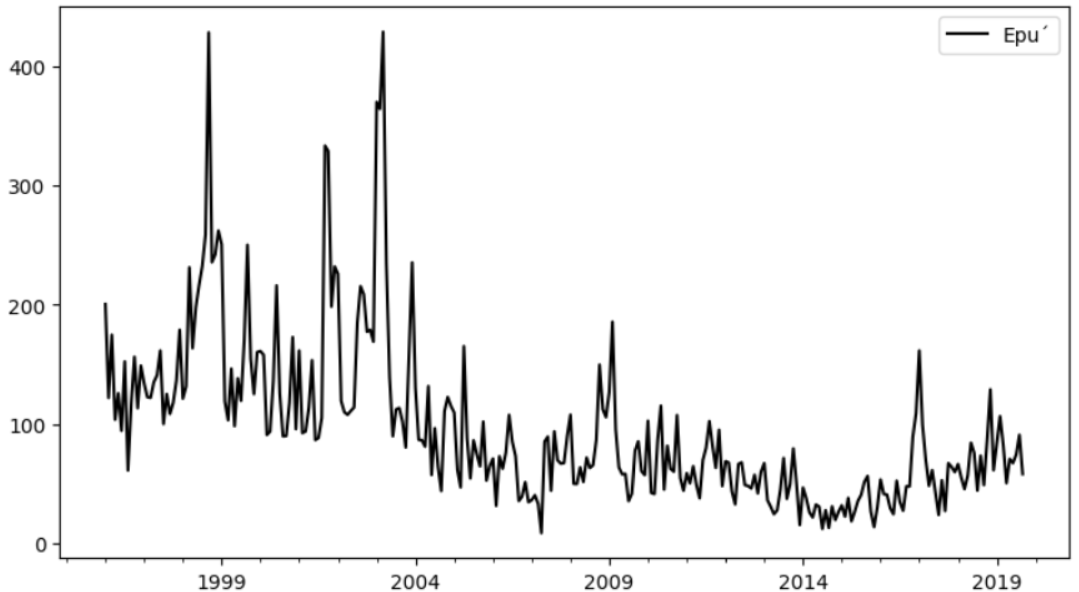}
    \vspace{1ex}
    \end{minipage}
    \begin{minipage}[b]{0.5\linewidth}
    \centering
    \subcaption{EPU-word2vec}
    \centering
    \includegraphics[width=.8\linewidth]{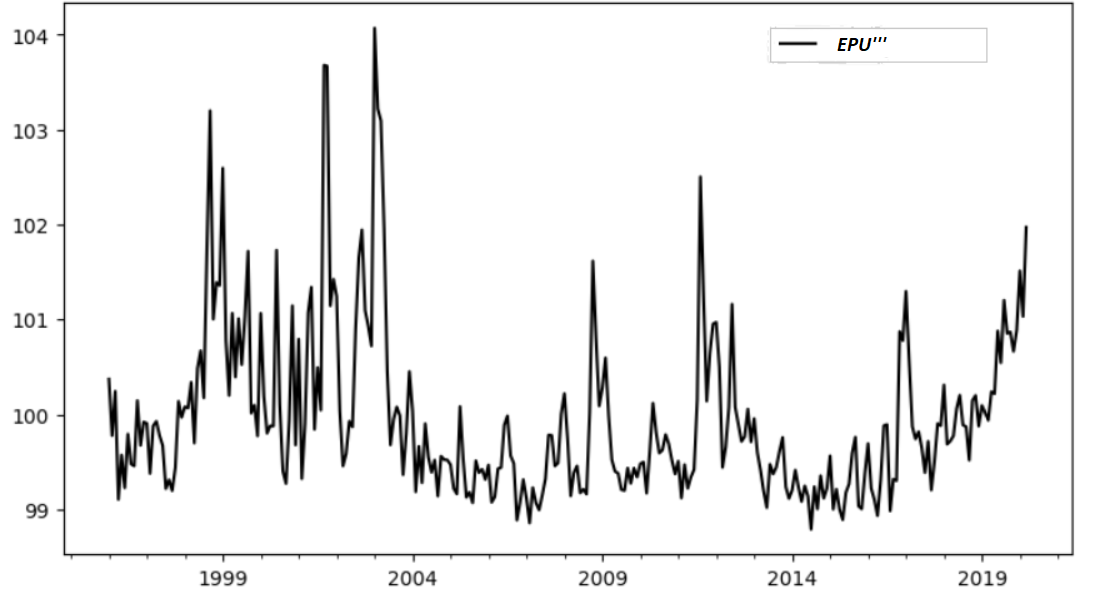}
    \vspace{1ex}
    \end{minipage}
    \caption{EPU index for Mexico, from January 1996 to December 2019}    
    \label{fig:chap3.fig29a}
\end{figure}

In the case of Mexico, we only compare the EPU index with Baker's approach because the index has not been calculated by Azqueta. Figure \ref{fig:chap3.fig29a} shows Baker's EPU index for Mexico and the index obtained with our proposal, with the sample expanded to cover 2019. Both indexes capture the 1993 and 2002 economic uncertainty crises, the 2008 global economic recession is more remarkable in the EPU-word2vec,  and the uncertainty measured by EPU-word2vec is more volatile during the 2012 political election and the presidency of Andres Manuel López Obrador in 2018.  Furthermore, the EPU-word2vec index reflects the uncertainty due to the recent COVID-19 pandemic that began at the beginning of 2020.

The DTW statistic between EPU-Baker and EPU-word2vec is $6.990$, and the correlation value is $0.8109$. In this case DTW value is just referential, because we just have Baker's index to make the comparison, but we can observe that this value is not very different from those we obtained for the US EPU index. Also, we can assume a linear correlation between the indexes, given that the correlation coefficient is $0.8109$. 

\subsection{Complement index} 
\label{Complement index}

\begin{figure}[h] 
 \begin{minipage}[b]{0.3\linewidth}
    \centering
    \includegraphics[width=1.1\linewidth]{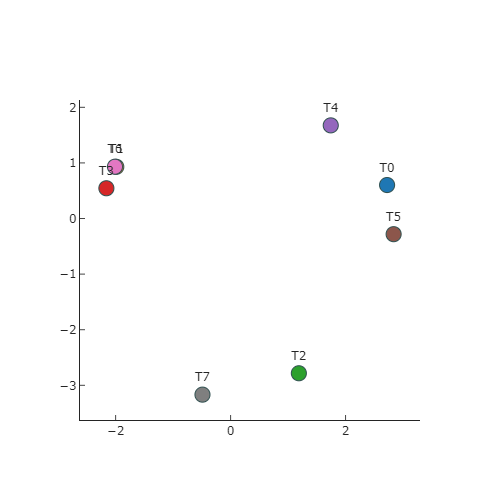}
    \subcaption{word2vec} 
    \end{minipage}
    \begin{minipage}[b]{0.3\linewidth}
    \centering
  \includegraphics[width=1.1\linewidth]{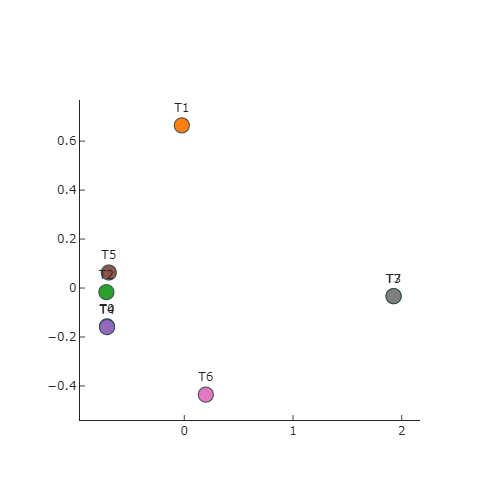}
  \subcaption{word2vec-pretrained} 
  \end{minipage}
   \begin{minipage}[b]{0.3\linewidth}
    \centering
  \includegraphics[width=1.1\linewidth]{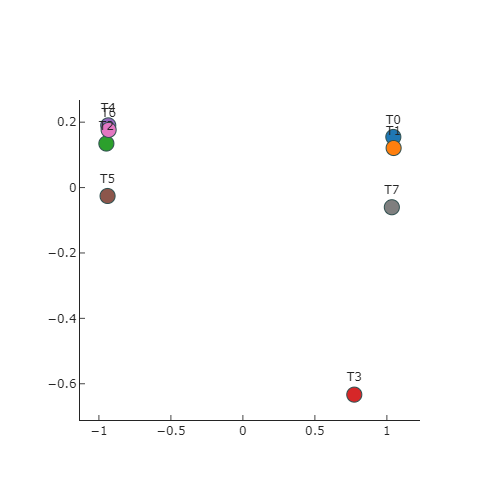}
  \subcaption{GloVe} 
  \end{minipage}
  \caption{Centroids of clusters (topics) for different configurations obtained with the proposed methodology and differents word embeddings for US news. We show the two first principal components of PCA.}
    \label{fig:chap3.fig19}
\end{figure}

\begin{figure}[h] 
   \begin{minipage}[b]{0.3\linewidth}
    \centering
    \includegraphics[width=1.1\linewidth]{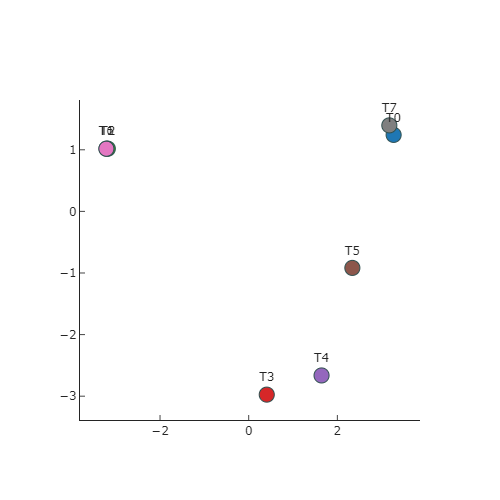}
    \subcaption{word2vec} 
    \end{minipage}
    \begin{minipage}[b]{0.3\linewidth}
    \centering
  \includegraphics[width=1.1\linewidth]{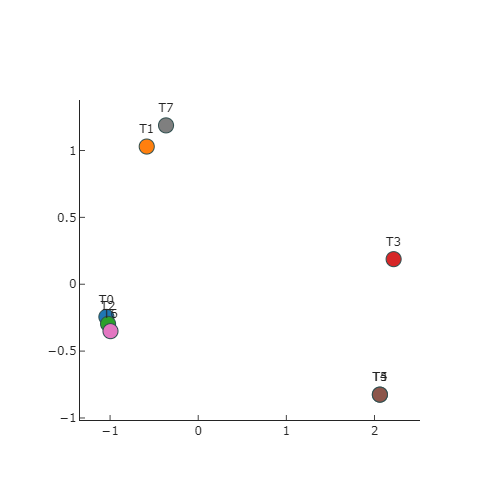}
  \subcaption{word2vec-pretrained} 
  \end{minipage}
   \begin{minipage}[b]{0.3\linewidth}
    \centering
  \includegraphics[width=1.1\linewidth]{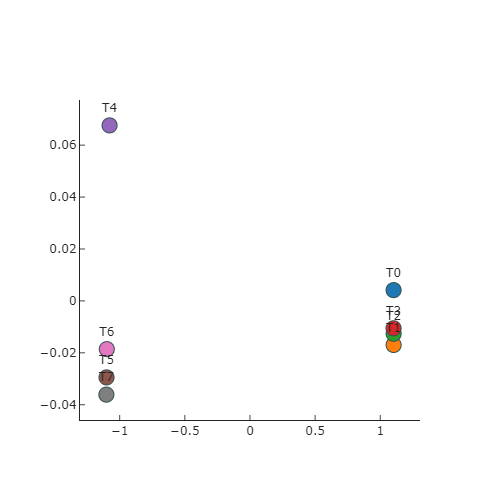}
  \subcaption{GloVe} 
  \end{minipage}
\caption{Centroids of clusters (topics) for different configurations obtained with the proposed methodology and differents word embeddings for Mexico news. We show the two first principal components of PCA}
    \label{fig:chap3.fig20}
\end{figure}

One of the advantages of the proposed method is the ability to identify groups on the EPU index subcategories using similarity in the semantic space spanned by the word embeddings. In this subsection, we describe our proposal to build a useful EPU index with a few subcategories. We will use the topic configuration obtained with word embeddings and the cosine distance to measure the similarity between topics represented by its centroids, which can be seen in Figures \ref{fig:chap3.fig19} and \ref{fig:chap3.fig20} for the US and Mexico news corpus, respectively.

\begin{figure}[ht] 
  \begin{subfigure}[b]{0.5\linewidth}
    \centering
    \includegraphics[width=0.7\linewidth]{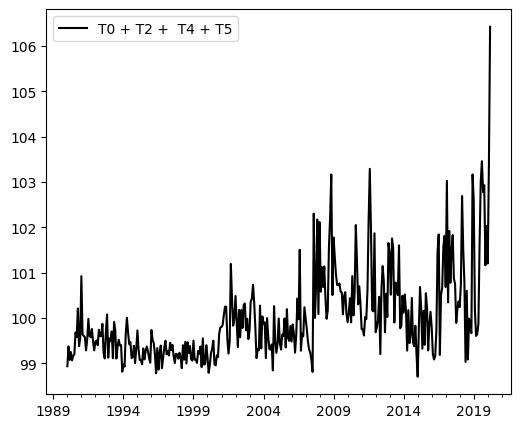}
    \caption{}
    \vspace{4ex}
  \end{subfigure}
  \begin{subfigure}[b]{0.5\linewidth}
    \centering
    \includegraphics[width=0.75\linewidth]{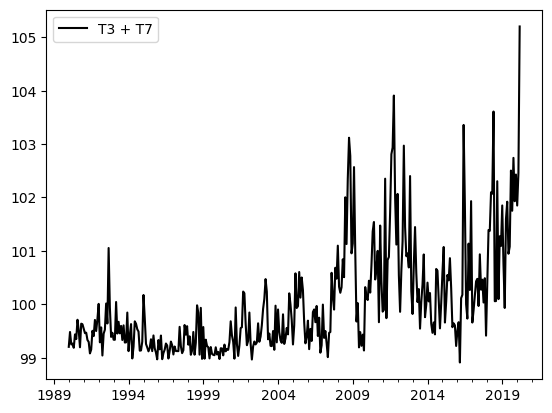} 
    \caption{} 
    \vspace{4ex}
  \end{subfigure} 
  \begin{subfigure}[b]{0.5\linewidth}
    \centering
    \includegraphics[width=0.75\linewidth]{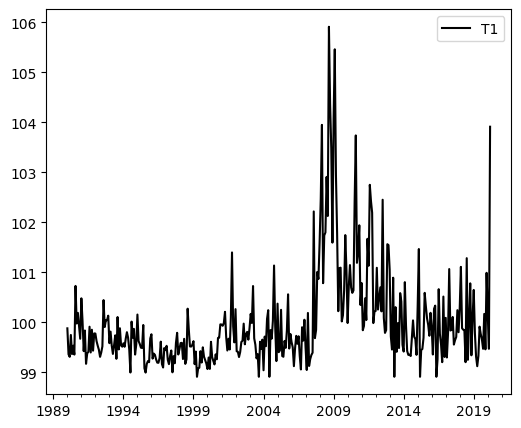} 
    \caption{} 
  \end{subfigure}
  \begin{subfigure}[b]{0.5\linewidth}
    \centering
    \includegraphics[width=0.75\linewidth]{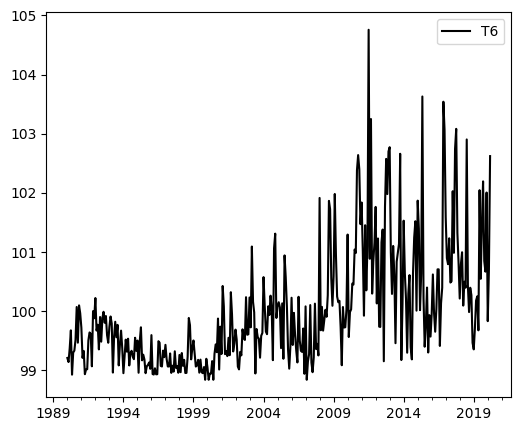} 
    \caption{} 
  \end{subfigure} 
  \begin{subfigure}[b]{0.5\linewidth}
    \centering
    \includegraphics[width=0.75\linewidth]{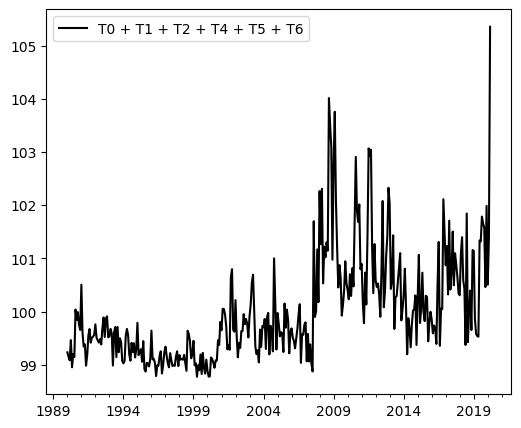} 
    \caption{} 
  \end{subfigure}
  
  \caption{US indexes (from January 1990 to March 2020)}
    \label{fig:chap3.fig30}
\end{figure}

Figure \ref{fig:chap3.fig30} presents the results for US news. The EPU index can be obtained just with topic 1, and according to Table \ref{tab:US EPU subcategories}, this topic is related to uncertainty in financial regulation and the Monetary-Fiscal and Trade Policy. Additionally, when we combine topic 0 (Stock market), topic 2 (Current crisis), topic 4 (Uncertainty), and topic 5 (Financial regulation), we obtain the uncertainty caused by financial regulations due to the crisis. By contrast, if we sum topic 3 (Current crisis) and topic 7 (National Security), we obtain the uncertainty related to national security caused by the current crisis. Furthermore, if we combine almost all the subcategories without taking into account the stock market and Fiscal-Monetary Policy, Figure \ref{fig:chap3.fig30}(e)  produces a less volatile EPU index during electoral periods, with a remarkable effect in the recent recession. 

\begin{figure}[ht] 
  \begin{subfigure}[b]{0.5\linewidth}
    \centering
    \includegraphics[width=0.75\linewidth]{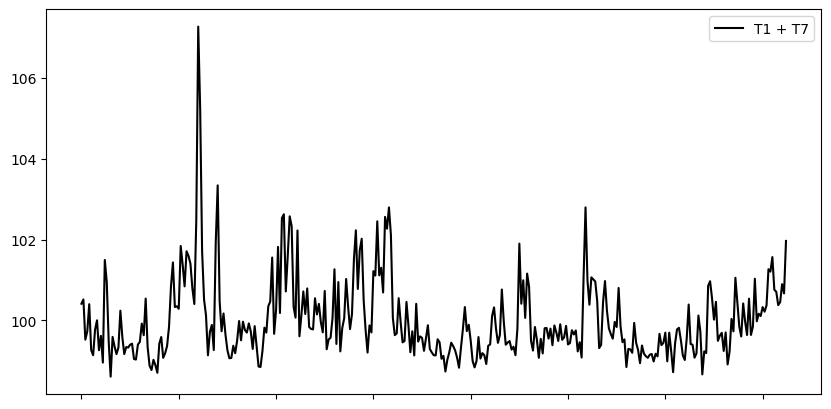}
    \caption{}
    \vspace{4ex}
  \end{subfigure}
  \begin{subfigure}[b]{0.5\linewidth}
    \centering
    \includegraphics[width=0.75\linewidth]{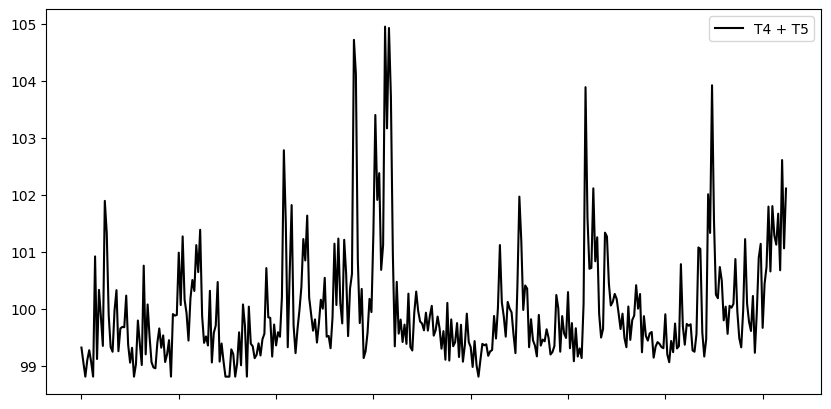} 
    \caption{} 
    \vspace{4ex}
  \end{subfigure} 
  \begin{subfigure}[b]{0.5\linewidth}
    \centering
    \includegraphics[width=0.75\linewidth]{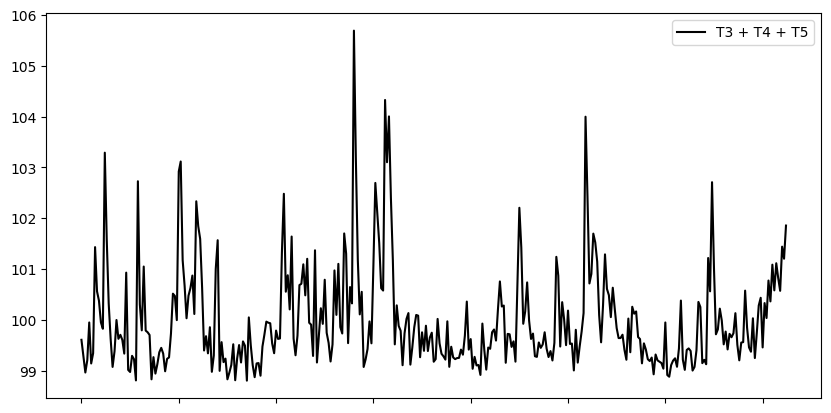} 
    \caption{} 
  \end{subfigure}
  \begin{subfigure}[b]{0.5\linewidth}
    \centering
    \includegraphics[width=0.75\linewidth]{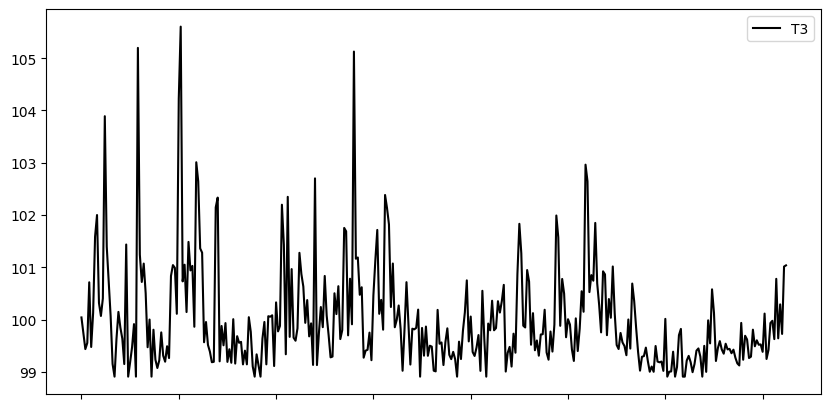} 
    \caption{} 
  \end{subfigure} 

  \caption{Mexico indexes (from January 1990 to March 2020)}
 \label{fig:chap3.fig31}
\end{figure}

Figure \ref{fig:chap3.fig31} presents the results for Mexico news. The EPU index can be recovered using topic 1, which is related to fiscal policy (see Table \ref{tab:MX EPU subcategories}) and topic 7 (fiscal-monetary policy and sovereign debt). If we combine topic 1 (fiscal policy) and topic 5 (trade policy), we obtain the uncertainty related to international regulation. However, we can combine topic 3 (stock market), topic 4 (global economy) and topic 5 (trade policy) to obtain an international economic uncertainty index.

\section{Conclusions}
\label{sec:conclusions}

In this paper, we proposed a methodology to obtain an index to measure economic uncertainty based on the information provided by news from digital media in  United States and Mexico in a similar way the EPU index does. Our proposal is based on an approximation to the results obtained by LDA in order to identify the latent topic structure in a vectorial representation of words and documents in the so called semantic space spanned by word embeddings, and using a similarity measure based on the cosine distance between words and documents in this representation. We explored different approaches for word and document embeddings, namely, word2vec and GloVe with and without transfer learning.

We made an extensive comparison of topic modeling with LDA and our proposal based on word embeddings and fuzzy $k-$means. The results shows that our proposal can approximate the main topic structure of news documents, but is fastest and computationally more efficient than LDA, particularly in the dynamic assignation of news documents into topics. 

Our results shows that we can reproduce the EPU index and most of its subcategories. Moreover, we can combine subcategories (topics) semantically related in order to build useful uncertainty indexes with interesting properties. We were able to obtain the EPU index where it was possible to identify significant events related to recent crises in both countries. 

As future work, we intend to compare the indicators using the probability distributions of the words and documents estimated by both methodologies and different word embeddings. Additionally, with the heuristic method, we will attempt to update the EPU index not only by date but also with changes in the topics according to the context.

\bibliographystyle{unsrtnat}
\bibliography{references}  

\end{document}